\documentclass[11pt]{article}

\usepackage[final]{acl}

\usepackage{times}
\usepackage{latexsym}
\usepackage{amsmath}
\usepackage{graphicx}
\usepackage{booktabs}
\usepackage{multirow} 
\usepackage{mathtools}
\usepackage{tabularx}
\usepackage{cancel}
\usepackage{amssymb}
\usepackage{xurl}
\usepackage{pifont}
\usepackage{booktabs}
\usepackage{colortbl}
\usepackage{tikz}
\usepackage{arydshln}
\usepackage{makecell}
\usepackage{algorithm}
\usepackage{algpseudocode}
\usepackage{tcolorbox}
\tcbuselibrary{listings, breakable, skins}
\usepackage{listings}
\usetikzlibrary{shapes, arrows.meta, positioning, fit, shadows, backgrounds, calc}
\usepackage{xcolor}
\newtcblisting{promptbox}[2][]{%
  colback=gray!5!white,
  colframe=gray!75!black,
  title={\textbf{#2}},
  fonttitle=\small\sffamily,
  breakable,
  enhanced,
  listing only,
  listing options={
    basicstyle=\small\ttfamily,
    breaklines=true,
    breakatwhitespace=true,
    columns=fullflexible,
    keepspaces=true,
    escapechar=|,
    aboveskip=0pt,
    belowskip=0pt,
    showstringspaces=false,
    literate={—}{---}1 {–}{--}1 {“}{``}1 {”}{''}1 {’}{'}1
  },
  left=5pt, right=5pt, top=5pt, bottom=5pt,
  #1
}
\newcommand{\placeholder}[1]{\textcolor{blue}{\{\{#1\}\}}}
\newtcolorbox{promptboxtext}[2][]{%
  colback=gray!5!white,
  colframe=gray!75!black,
  title={\textbf{#2}},
  fonttitle=\small\sffamily,
  breakable,
  enhanced,
  fontupper=\small, 
  boxrule=0.5pt,    
  left=5pt, right=5pt, top=5pt, bottom=5pt,
  #1
}
\usepackage[T1]{fontenc}

\usepackage[utf8]{inputenc}

\usepackage{microtype}

\usepackage{inconsolata}

\usepackage{graphicx}
\usepackage{enumitem}

\newcommand{\rc}[1]{#1}

\newcommand{\dy}[1]{#1}

\title{DR-Arena: an Automated Evaluation Framework for \\ Deep Research Agents}

\author{
    \textbf{Yiwen Gao}\textsuperscript{1}\thanks{\ \ Equal contribution.} \quad
    \textbf{Ruochen Zhao}\textsuperscript{2}\footnotemark[1] \quad
    \textbf{Yang Deng}\textsuperscript{3} \quad
    \textbf{Wenxuan Zhang}\textsuperscript{4}\thanks{\ \ Corresponding author.} \\
    \textsuperscript{1}National University of Singapore \quad
    \textsuperscript{2}Nanyang Technological University \\
    \textsuperscript{3}Singapore Management University \quad
    \textsuperscript{4}Singapore University of Technology and Design \\
    \texttt{yiwen\_gao@u.nus.edu} \quad \texttt{ruochen002@e.ntu.edu.sg} \\
    \texttt{ydeng@smu.edu.sg} \quad \texttt{wxzhang@sutd.edu.sg} \\ \\
    \url{https://github.com/iNLP-Lab/DR-Arena}
}

\begin{document}
\maketitle
\begin{abstract}
\dy{As Large Language Models (LLMs) increasingly operate as \textbf{Deep Research (DR) Agents} capable of autonomous investigation and information synthesis, reliable evaluation of their task performance has become a critical bottleneck.}
Current benchmarks predominantly rely on static datasets, which suffer from 
\dy{several limitations: \textit{limited task generality}, \textit{temporal misalignment}, and \textit{data contamination}.}
To address these, we introduce \textbf{DR-Arena}, a fully automated evaluation framework that pushes 
\dy{DR} agents to their \dy{capability}
limits through 
\dy{dynamic investigation}. DR-Arena constructs real-time Information Trees from fresh web trends to ensure the evaluation rubric is synchronized with the live world state, and employs an automated Examiner to generate structured tasks testing two orthogonal capabilities:  \textit{Deep reasoning} and \textit{Wide coverage}. 
\dy{DR-Arena further adopts} Adaptive Evolvement Loop, a state-machine controller that dynamically escalates task complexity based on real-time performance, demanding deeper deduction or wider aggregation until a decisive capability boundary emerges. 
Experiments with six advanced 
\dy{DR} agents demonstrate that DR-Arena achieves a Spearman correlation of {0.94} with the LMSYS Search Arena leaderboard. This represents state-of-the-art alignment with human preferences without any manual efforts, validating DR-Arena as a reliable alternative for costly human adjudication. 
\end{abstract}

\section{Introduction}

Deep Research (DR) agents, such as OpenAI Deep Research \citep{openai2025deepresearch} and Perplexity \rc{Deep Research} \citep{perplexity2025}, have rapidly gained adoption and are now widely used for complex information-seeking tasks. Unlike traditional search engines where users need to browse multiple websites manually, DR agents act as autonomous research agents that conduct multi-step investigations over extended horizons, iteratively retrieving, cross-referencing, and synthesizing evidence from the live web to produce structured and citation-backed reports \citep{nakano2021webgpt, you2024storm, qin2023webcpm}. As \rc{DR} agents are increasingly deployed in real-world and high-stakes analytical settings, \rc{efficient} and reliable evaluation of their capabilities has become a pressing challenge.

Recent efforts have begun to 
evaluate deep research 
capabilities by constructing dedicated datasets \rc{for} 
multi-step web-based investigation \citep{mialon2023gaia, wong2025widesearch, lan2025deepwidesearch}. While these benchmarks provide useful performance indicators, they 
\dy{exhibit three fundamental limitations: 1) \textbf{limited task generality}, as task-centric and aspect-specific dataset construction restricts evaluation to predefined investigation patterns and weakens transferability to real-world research settings; 2) \textbf{temporal misalignment}, since static benchmarks inevitably decay as underlying facts evolve, resulting in evaluation against outdated ground truth; and 3) \textbf{data contamination}, whereby persistent and widely reused datasets increasingly appear in model training corpora, resulting in parametric memorization rather than genuine reasoning and evidence synthesis \citep{ravaut2024much}.}
Collectively, these issues reflect a structural mismatch with DR agents, which are designed to operate in open evolving environments that require adaptive exploration and reasoning. 

\begin{figure*}[t]
    \setlength{\abovecaptionskip}{3pt} 
    \setlength{\belowcaptionskip}{0pt}
    \centering
    \includegraphics[width=\textwidth]{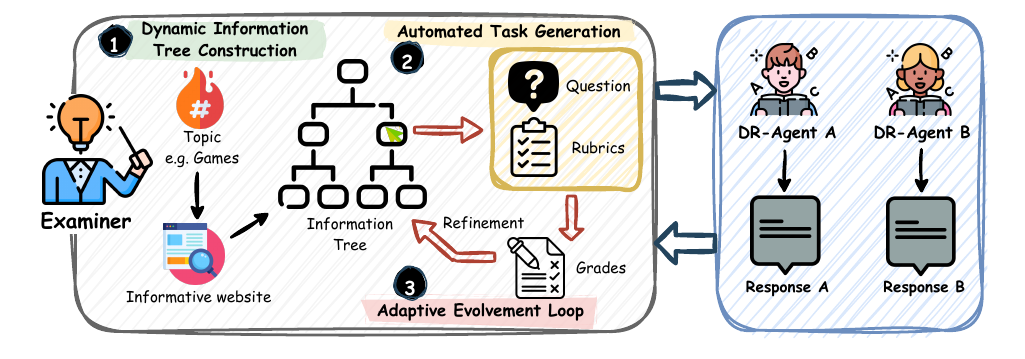}
    \caption{Overview of the DR-Arena Framework.}
    \label{fig:framework}
    \vspace{-3mm}
\end{figure*}

To address these challenges, we propose DR-Arena, a fully automated evaluation framework 
\rc{that simulates a competitive arena, where DR agents are pushed to their 
\dy{capability} limits through dynamic investigation.} As shown in Figure \ref{fig:framework}, 
DR-Arena operates as a closed-loop framework orchestrated by an automated Examiner\rc{, which functions as an interviewer role that drafts questions, assesses candidate answers, and conducts follow-up rounds}. 
\rc{To draft dynamic questions, the Examiner first constructs Information Trees by scraping informative websites in real-time. Based on the trees, the Examiner then devises a challenging question for the candidate DR agents. Since the Examiner has access to the ground-truth information source, it also keeps a mental set of grounded rubrics. After the agents respond, the Examiner assesses their performances based on the rubrics. Finally, it enters the Adaptive Evolvement Loop, where the Examiner dynamically decides whether to further refine the tree and ask follow-up questions.}

\rc{During Adaptive Evolvement, DR-Arena stress-tests two fundamental dimensions: multi-hop reasoning (\textit{Depth}) and information gathering (\textit{Width}). The Examiner actively intervenes based on real-time performance to amplify differences between candidates: When agents reach a stalemate, the Examiner first diagnoses the failure type, whether the mistake originates from a retrieval gap or reasoning deficit, and then intentionally probes that specific weakness in subsequent rounds. This dynamic investigation, guided by the 
\dy{\textit{Depth} and \textit{Width} principles \citep{lan2025deepwidesearch}}, ensures that both core abilities are tested comprehensively, pushing DR agents to their capability boundaries.}


Experiments across six state-of-the-art DR agents \rc{show} 
that DR-Arena achieves superior alignment with human preferences compared to existing benchmarks, \rc{accurately} recovering the hierarchy of the LMSYS Search Arena 
\dy{\cite{miroyan2025search}} \rc{even} without manual annotations. Beyond alignment, we demonstrate that the adaptive loop optimizes computational allocation, concentrating computational resources on distinguishing closely matched models while rapidly resolving mismatches. Furthermore, \rc{we conduct extensive}
analysis \rc{that} reveals distinct cognitive profiles among top-tier \rc{DR} agents, distinguishing models with robust, balanced capabilities from those exhibiting asymmetric trade-offs between logic and coverage. 
\rc{In conclusion,} DR-Arena \rc{serves as} 
a trustworthy and scalable proxy for human adjudication, offering the community a reliable standard to benchmark the next generation of \rc{DR} 
agents.

\section{Related Work}
\label{sec:related_work}

\paragraph{Static Benchmarks for Deep Research.}
Evaluation in Agentic Search has traditionally relied on {static benchmarks}. Early works like \textit{WebArena} \citep{zhou2023webarena} and \textit{BrowseComp} \citep{deng2024browsecomp} focused on transactional web tasks\dy{, while later benchmarks extended to deep research via expert-curated datasets, including \textit{DeepResearch Bench} \citep{du2025deepresearchbench} and \textit{DeepResearch Arena} \citep{wan2025deepresearcharena}.} 
\dy{Despite increased task complexity, these benchmarks remain static and thus suffer from \textit{temporal degradation} \citep{kasai2024realtimeqa}: As real-world information evolves, fixed gold answers become outdated, penalizing agents that correctly retrieve fresh evidence \citep{vu2023freshllms}. Approaches such as \textit{LiveBench} \citep{white2025livebench} mitigate contamination through periodic updates, but still rely on batch refreshes rather than real-time generation.}
\dy{DR-Arena fundamentally departs from this paradigm by generating Dynamic Information Trees directly from the live web, synchronizing tasks and ground truth with the current world state and eliminating both temporal misalignment and parametric memorization.}

\paragraph{Automated Evaluation Frameworks.}
\dy{To scale beyond human review, prior work has adopted automated evaluation frameworks, most notably \textit{LLM-as-a-Judge} \citep{zheng2023judging} and its refinements in leaderboards such as \textit{AlpacaEval} \citep{li2023alpacaeval} and \textit{Auto-Arena} \citep{zhao2024autoarena}, which reduce single-judge bias via multi-agent deliberation with ``Peer Battle'' and ``Committee'' mechanisms.} 
In the agentic domain, frameworks like \textit{Mind2Web-2} \citep{gou2025mind2web2} and \textit{Auto-Eval Judge} \citep{bhonsle2025autoeval} have evolved into ``Agent-as-a-Judge'' systems, utilizing structured checklists to verify step-by-step execution.
However, the majority of these frameworks operate as {passive evaluators}, where they grade static trajectories after the task is completed. 
\dy{While \textit{FACT-AUDIT} \citep{lin2025fact} introduces adaptive stress-testing, it remains limited to short-horizon factual claims.}
\dy{DR-Arena departs from this paradigm by introducing an active Examiner for long-horizon Deep Research, where an \textit{Adaptive Evolvement Loop} dynamically modulates task \textit{depth} and \textit{width} during evaluation to expose capability boundaries that static datasets or passive evaluators fail to expose.}

\section{DR-Arena Framework}
\label{sec:framework}
\begin{figure*}[t]
    \setlength{\abovecaptionskip}{3pt} 
    \setlength{\belowcaptionskip}{0pt}
    \centering
    \includegraphics[width=\textwidth]{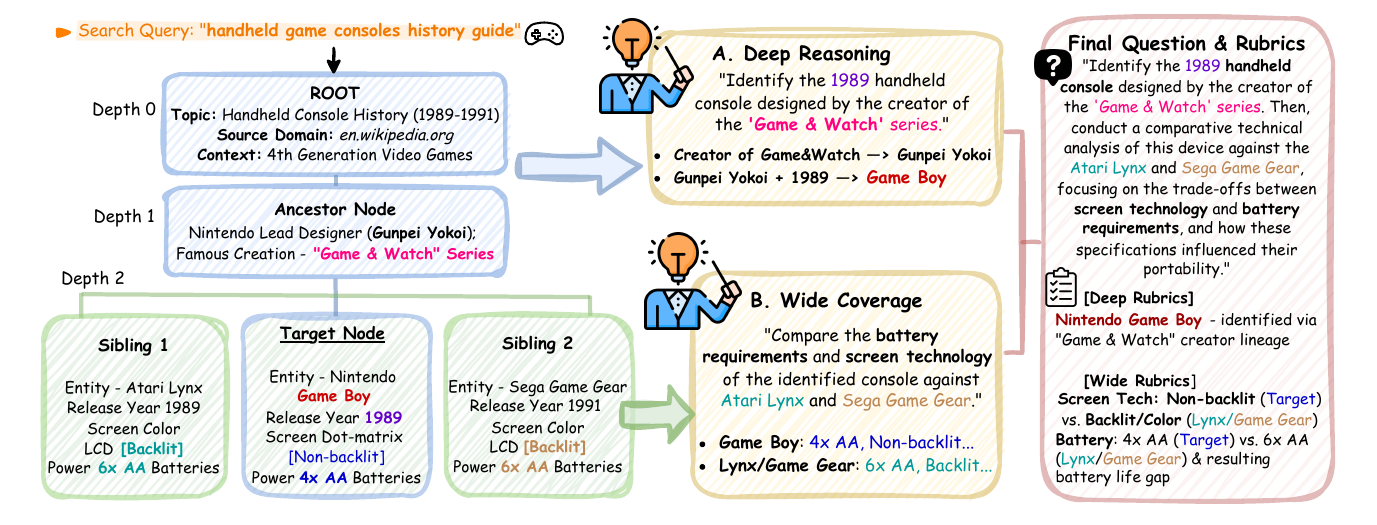}
    
    \caption{Automated Task Generation via Dynamic Information Trees.}
    
    \label{fig:tree_mapping}
    \vspace{-3mm}
\end{figure*}

To probe the limits of DR agents, DR-Arena operates as a closed-loop system orchestrated by a unified LLM agent, denoted as {Examiner}. As illustrated in Figure \ref{fig:framework}, the pipeline consists of three sequential stages: Dynamic Information Tree Construction, Automated Task Generation, and Adaptive Evolvement Loop. Detailed prompts and configuration settings are provided in Appendix \ref{app:prompts}. 

\subsection{Dynamic Information Tree Construction}
\label{subsec:tree_construction}
To ensure that the evaluation reflects the complexity and timeliness of the real internet, 
\rc{DR-Arena} 
directly scrapes and structures real-world websites into a verifiable knowledge base, as illustrated by the instantiated example in Figure \ref{fig:tree_mapping}. 

\rc{To ensure task diversity, the Examiner first samples a seed topic from Google Trends\footnote{\url{https://github.com/pat310/google-trends-api/wiki/Google-Trends-Categories}} and navigates to a specific sub-topic, e.g. \textit{``Handheld Consoles''} $\rightarrow$ \textit{``Nintendo History''}. Then, it generates a relevant search query to retrieve top websites from the open web. Among the top returned URLs, the Examiner selects a high-quality informative website to use as a ``Root Node'' ($v_{root}$) for the Information Tree, which is typically a comprehensive guide or hub page with rich outbound links}

From $v_{root}$,
the system \rc{constructs an Information Tree by scraping} 
this page and its linked neighbors.
\rc{We could represent the Information Tree as}
a directed graph $G = (V, E)$.
Each node $v_i \in V$ represents a real webpage, containing the full text content $C_i$ and metadata (URL, title). An edge \rc{$e_{ij} \in E$} exists if node $v_i$ contains a hyperlink to $v_j$. 
\rc{To capture inter-node relationships $R_{ij}$, we enrich edges with semantic context by analyzing anchor text and surrounding content to derive relationship types such as varieties or processes.}

\rc{While we limit the Information Tree size at first, the tree can be expanded in two orthogonal directions during the process:}
Depth Expansion recursively crawls specific branches to depth $d$ to establish reasoning chains (e.g., Handheld Console History $\rightarrow$ Specific Console $\rightarrow$ Technical Specs) supporting \textit{Multi-hop Reasoning Tasks}; Width Expansion captures sibling nodes sharing the same parent and semantic relationship (e.g., multiple handheld consoles listed on the history page) to support aggregation tasks. 

\subsection{Automated Task Generation}
\label{subsec:task_generation}

Leveraging the \rc{Information Tree},
the Examiner dynamically generates questions to probe two core capabilities of DR agents: multi-hop reasoning (\textit{depth}) and information gathering (\textit{width}). 

\paragraph{\rc{
\dy{Depth and Width} Principles.}}
When generating questions, we utilize the definition of \textit{depth} and \textit{width} 
\rc{from} \citet{lan2025deepwidesearch}. First, depth refers to the search steps required,
which could be directly visualized by the \rc{tree depth}. For example, 
\rc{deep reasoning tasks mask the target entity's name, requiring multi-hop deduction through relationship chains (e.g., ``Identify the 1989 handheld console designed by the creator of the Game \& Watch series'').}
\rc{Second, width refers to the number of information units to be searched, which can be visualized as tree width. Wide coverage tasks require synthesizing attributes across sibling nodes, demanding comprehensive aggregation through parallel entity comparisons (e.g., ``Compare the battery specifications of the identified console against its market competitors'').}
\rc{Collectively, \textit{depth} and \textit{width} constitute two fundamental capabilities of DR agents: reasoning and information gathering.}


\paragraph{\rc{Task Generation.}}
\rc{The Examiner then generates a challenging question based on the Information Tree. First, it randomly selects a node from the tree and performs a validation check:}
if the current node lacks sufficient depth (no children) or width (no siblings) to support a challenging query, the system automatically triggers the crawler to fetch fresh data, expanding the tree structure on-the-fly. 
\rc{Then, the Examiner drafts a challenging query to probe both depth and width capabilities. To prevent candidates from taking shortcuts,}
the Examiner employs a de-contextualization strategy:\rc{ When generating queries,} the Examiner is strictly prohibited from mentioning specific filenames \rc{or} website titles.

Since the ground truth ranging from logical lineages to attribute values is structurally encoded in the \rc{tree}, the \rc{Examiner simultaneously} 
synthesizes a \rc{list of rubrics that is}
factually correct and traceable to source URLs. This \rc{list} 
includes a \textit{Checklist-Depth} for verifying logical identity and a \textit{Checklist-Width} for assessing data completeness. Representative task examples are provided in Appendix \ref{app:example_tasks}.

\subsection{\rc{Adaptive Evolvement Loop}}

\rc{After task generation, the Examiner assesses the candidate responses against the rubrics and enters the Adaptive Evolvement Loop, where it dynamically decides whether to draft follow-up questions to push candidates to their capability boundaries and hence make the performance gaps more visible.}

\begin{table}[h]
\centering
\small
\renewcommand{\arraystretch}{1.4}
\begin{tabularx}{\linewidth}{@{} l X @{}}
\toprule
\textbf{Failure Tag} & \textbf{Diagnostic Criteria} \\
\midrule
\textbf{\textsc{Depth}} & \textbf{Logic Failure.} Failed to identify the correct core entity due to a broken reasoning chain. \\
\textbf{\textsc{Width}} & \textbf{Coverage Failure.} Failed to aggregate specific attribute details (Data Gap). \\
\textbf{\textsc{Both}} & \textbf{Systemic Failure.} Failed on both logical identification and factual completeness. \\
\textbf{\textsc{None}} & \textbf{Soft Gap.} The loss was determined solely by soft filters (formatting or utility preferences). \\
\bottomrule
\end{tabularx}
\caption{\textbf{Taxonomy of Failure Types.} The Examiner provides a diagnostic tag \rc{in verdicts.} 
}
\label{tab:failure_types}
\vspace{-3mm}
\end{table}

\begin{table*}[t]
\centering
\resizebox{\textwidth}{!}{
\begin{tabular}{l l l l}
\toprule
\textbf{Adjudication Verdict} & \textbf{Diagnostic Signal} & \textbf{Evolution Action} & \textbf{Strategic Rationale} \\
\midrule
\textbf{Tie} (High Quality) & N/A & \textbf{Pressure Test} ($D \uparrow 1$ \& $W \uparrow 1$) & Current task too easy; find ceiling. \\
\hline
\textbf{Tie} (Low Quality) & N/A & \textbf{Backtrack} (Move to Parent) & Current task too hard; re-establish baseline. \\
\midrule
\multirow{3}{*}{\textbf{Winner Decided}} & \textsc{Depth} (Logic Failure) & \textbf{Probe Depth} ($D \uparrow 1$) & Challenge loser's reasoning capabilities. \\
\cline{2-4}
& \textsc{Width} (Coverage Failure) & \textbf{Probe Width} ($W \uparrow 1$) & Challenge loser's information coverage. \\
\cline{2-4}
& \textsc{Both} / \textsc{None} & \textbf{Pressure Test} ($D \uparrow 1$ \& $W \uparrow 1$) & Ambiguous failure; increase difficulty. \\
\bottomrule
\end{tabular}
}
\caption{\textbf{The Evolvement Loop Transition Matrix.} The system dynamically adjusts task complexity (Depth $D$ and Width $W$) based on adjudication verdict and diagnostic failure type of the losing agent.}
\label{tab:evolvement_logic}
\vspace{-3mm}
\end{table*}


\paragraph{\rc{Evidence-Based Judgment.}}
Evaluating open-ended \rc{reports}
is inherently challenging due to the lack of ground-truth data. DR-Arena addresses this by employing the Examiner as a Judge, utilizing the generated rubrics via a strict two-stage protocol: First, the \rc{Examiner}
verifies \textit{Hard Constraints} by checking \rc{against the list of rubrics}.
A critical error, such as misidentifying the core entity (logic error) or omitting mandatory data points (coverage gap), results in immediate penalties. Then, the \rc{Examiner}
evaluates \textit{Soft Constraints}, comparing user experience \rc{aspects}
such as presentation quality, formatting, information density, and helpfulness.

Instead of a simple binary win/loss, we adopt a tiered adjudication system to capture the magnitude of performance gaps \rc{with finer granularity. When a candidate (A or B) wins, the verdict distinguishes whether it is a decisive win or a marginal win. Ties are also}
classified into High Quality (mutual success) or Low Quality (mutual hallucination) to inform the subsequent evolution strategy.
\begin{equation}
\small 
\text{Verdict}(A, B) \in
\left\{
\begin{array}{@{}l@{}} 
  \textbf{\textsc{Much Better}} \text{ (Decisive Win)} \\[2pt]
  \textbf{\textsc{Better}} \text{ (Marginal Win)} \\[2pt]
  \textbf{\textsc{Tie}}: \begin{cases} 
       \text{Both High Quality} \\
       \text{Both Low Quality}
    \end{cases}
\end{array}
\right.
\end{equation}


\rc{Guided by the Depth and Width principles, the Examiner also performs a failure type analysis on the losing side's response: Whether it is a Depth failure, a Width failure, or both/none. Detailed explanations on these failure types are outlined in Table \ref{tab:failure_types}.}
This distinction is critical as it disentangles reasoning deficits from retrieval gaps, providing the necessary signal for the Evolvement Loop to intelligently target the specific weakness in the subsequent rounds.

\paragraph{\rc{Follow-up Rounds.}}
After each round, the \rc{Examiner}
analyzes the verdict and failure type to operate on a targeted probing strategy \rc{to accelerate performance divergence and reach a decisive verdict,} as outlined in Table \ref{tab:evolvement_logic}, 
if a \rc{High-quality} Tie
occurs, the system triggers a pressure test, increasing both depth ($D$) and width ($W$) parameters to locate the capability ceiling. Conversely, if a \rc{marginal} win \rc{occurs,}
the system aggressively targets the loser's specific weakness\rc{, probing either depth or width.}

This adaptive mechanism ensures the system efficiently converges to a verdict by continuously pushing agents toward their specific breakdown points. The loop terminates when a decisive ``Much Better'' verdict is reached, the cumulative score difference exceeds a threshold, or the maximum round limit is met. Complete pseudo-code and a full match walkthrough are in Appendix \ref{app:algorithm} and \ref{app:full_match}.

\section{Experiments}
\label{sec:experiments}


\subsection{Experimental Setup}

\paragraph{\rc{Model Choice.}}
From the 
pool of 13 models listed on LMSYS Search Arena 
(as of Dec 3, 2025), we select 6 representative models covering the latest or best-performing versions from major model families:
GPT-5.1-Search (OpenAI),
Gemini-2.5-Pro-Grounding (Google),
o3-Search (OpenAI),
Grok-4-Search (xAI),
Perplexity-Sonar-Pro-High (Perplexity), and Claude-Opus-4.1-Search (Anthropic). As good-performing LLMs tend to perform better in LLM-as-a-judge tasks~\citep{zheng2023judging}, we choose Gemini-3-Pro as the fixed Examiner, \rc{which} 
is currently the best model 
on LMSYS \textit{Text} Arena.

\paragraph{\rc{Baselines.}}
\dy{We compare DR-Arena against six recent benchmarks for evaluating DR search agents, including BrowseComp \citep{deng2024browsecomp}, DeepResearch Bench \citep{du2025deepresearchbench}, LiveNewsBench \citep{li2026livenews}, LiveSearchBench \citep{zhou2025livesearch}, LiveResearchBench \citep{wang2025liveresearch}, and Deep Research Bench \citep{bosse2025DRsearch}. Details of these benchmarks are shown in Appendix \ref{app:benchmarks}.}
As a reference, we utilize the LMSYS Search Arena scores as the human-annotated ground truth for evaluating alignment of the resulting rankings. They rely on large-scale blind human comparisons to establish trustworthy model rankings.

\paragraph{Tournament Configuration.}
\rc{To maximize computational efficiency, we conduct a Swiss-style tournament: For $n$ participants, instead of pairing each participant with $(n-1)$ others, a Swiss tournament pairs each player with $\lceil \log_2(n) \rceil$ players of similar rankings without repeats. This design effectively reduces computational costs of ranking $n$ models from $O(n^2)$ to $O(n\log_2(n))$.}
To mitigate 
bias, the tournament begins with a randomized initialization, followed by 4 rounds of dynamic pairing. 
While the minimum number of rounds recommended by the Swiss tournament design is $\lceil \log_2 6 \rceil = 3$, we deliberately oversample slightly to reduce variance from early-stage stochasticity. To ensure both domain robustness and fair comparison, each pairing competes across the same set of 30 pre-constructed Information Trees derived from diverse real-world websites {(refer to Appendix \ref{app:dataset_stats} for detailed topic and domain distributions)}. 

\paragraph{Scoring Mechanism.}
Following competitive gaming standards \citep{elo1978rating}, we utilize the Elo rating system updated via the Bradley-Terry (BT) model \citep{bradley1952rank}. Elo scores are dynamically updated after each round to guide the subsequent pairings, ensuring that models continuously face opponents of comparable strength.

\subsection{Main Results: Alignment with Humans}
\label{subsec:main_results}
We evaluate the reliability of DR-Arena by benchmarking its derived rankings against the human-verified LMSYS Search Arena \rc{scores}.

\begin{table}[t]
\centering
\resizebox{\columnwidth}{!}{
\begin{tabular}{l cc c cc}
\toprule
\multirow{2}{*}{\textbf{Model}} & \multicolumn{2}{c}{\textbf{Search Arena}} & \phantom{a} & \multicolumn{2}{c}{\textbf{DR-Arena}} \\
\cmidrule{2-3} \cmidrule{5-6} 
& \textbf{Elo} & \textbf{Rank} && \textbf{Elo} & \textbf{Rank} \\
\midrule
GPT-5.1-Search             & 1201 & \textbf{1} && 1084 & \textbf{1} \\
Gemini-2.5-Pro-Grounding   & 1142 & \textbf{2} && 1054 & \textbf{2} \\
o3-Search                  & 1139 & \textbf{3} && 1041 & \textbf{3} \\
Grok-4-Search              & 1138 & \textbf{4} && 958  & \textbf{4} \\
Claude-Opus-4.1-Search     & 1130 & 5          && 921  & 6 \textcolor{red}{($\downarrow$)} \\
Perplexity-Sonar-Pro-High  & 1125 & 6          && 942  & 5 \textcolor{green}{($\uparrow$)} \\
\bottomrule
\end{tabular}
}
\caption{Comparison between DR-Arena Leaderboard and LMSYS Search Arena.} 
\label{tab:main_leaderboard}
\vspace{-3mm}
\end{table}



\begin{figure}[t]
    \setlength{\abovecaptionskip}{0pt} 
    \setlength{\belowcaptionskip}{0pt}
\centering
\includegraphics[width=0.95\columnwidth]{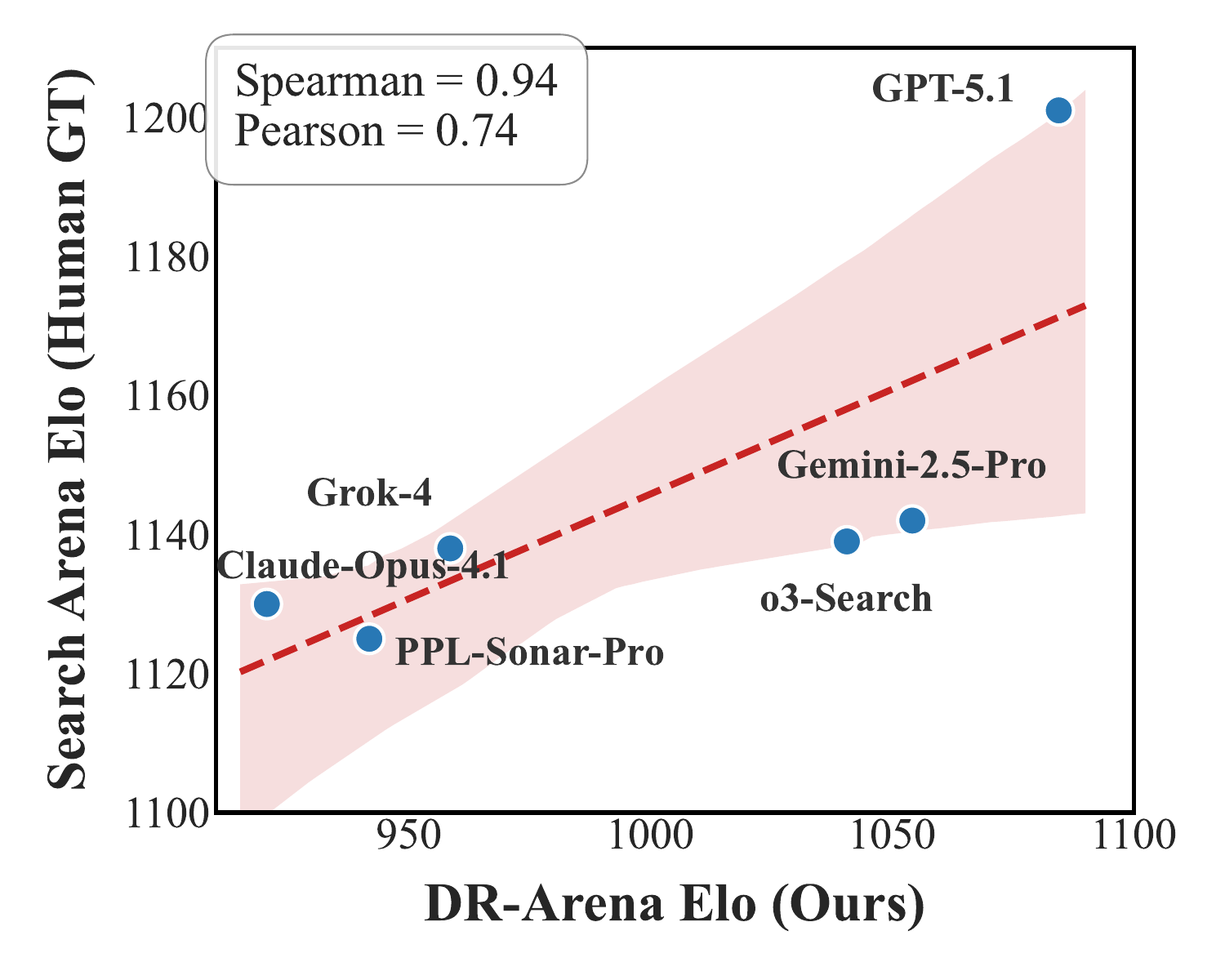}
\caption{Leaderboard Correlation Analysis. }
\label{fig:elo_scatter}
\vspace{-3mm}
\end{figure}

\paragraph{Leaderboard Consistency.}
Table \ref{tab:main_leaderboard} presents the comparative rankings derived from DR-Arena and \rc{LMSYS} Search Arena. DR-Arena achieves approximate alignment with human judgment, recovering the exact hierarchy for most models. Figure \ref{fig:elo_scatter} plots Elo scores derived from Search Arena against those derived from DR-Arena. We observe a 
linear mapping, confirming that our automated metrics effectively proxy human preference. 

While the overall alignment is high, a minor rank swap occurs between Perplexity-Sonar-Pro (\#5 in DR-Arena) and Claude-Opus-4.1 (\#6). As detailed in {Appendix \ref{app:qualitative_analysis}}, this divergence \rc{could be an example of}
the \textit{Factuality-Fluency Trade-off}: Our qualitative analysis reveals that the automated Examiner strictly penalizes hallucinated numbers that human annotators may overlook, being swayed by Claude's superior writing style.


\begin{table}[t]
\centering
\renewcommand{\arraystretch}{1.4} 
\setlength{\tabcolsep}{4pt}       

\resizebox{\columnwidth}{!}{
\begin{tabular}{l c c c}
\toprule
\textbf{Benchmark} & \textbf{N} & \textbf{Pearson} & \textbf{Spearman} \\
\midrule
\rowcolor{gray!15} \textbf{DR-Arena (Ours)} & \textbf{6} & \textbf{0.74} & \textbf{0.94} \\

BrowseComp \makecell[l]{\citep{deng2024browsecomp}} & 5 & 0.70 & 0.76 \\

DeepResearch Bench \makecell[l]{\citep{du2025deepresearchbench}} & 4 & 0.63 & 0.40 \\

LiveNewsBench \makecell[l]{\citep{li2026livenews}} & 4 & 0.46 & 0.20 \\

LiveSearchBench \makecell[l]{\citep{zhou2025livesearch}} & 4 & -0.23 & -0.11 \\

LiveResearchBench \makecell[l]{\citep{wang2025liveresearch}} & 4 & -0.47 & -0.63 \\

Deep Research Bench \makecell[l]{\citep{bosse2025DRsearch}} & 6 & -0.86 & -0.90 \\
\bottomrule
\end{tabular}
}
\caption{\textbf{Comparison with SOTA Benchmarks. } \rc{N denotes the number of models used for calculation.}}
\label{tab:benchmark_comparison}
\vspace{-3mm}
\end{table}

\paragraph{Superiority over Static Benchmarks.}
\rc{We also calculate the correlation scores between LMSYS Search Arena and other benchmark datasets. Results are shown in Table \ref{tab:benchmark_comparison}.}
We observe that DR-Arena achieves \rc{the SOTA} Spearman correlation of 0.94, and a Pearson correlation of 0.74, significantly outperforming static datasets like Deep Research Bench (FutureSearch) and LiveResearchBench, which show negative correlations. Notably, among all benchmarks, DR-Arena is the only one that requires zero human intervention, automating the ground-truth generation via Dynamic Information Trees to assess factual correctness on the live web. We \rc{hypothesize}
that this high alignment with humans stems from our \textit{Dynamic Investigation} process, which simulates the iterative nature of human research, capturing nuances like error recovery and synthesis that static QA metrics \rc{could} miss.


\subsection{Ablation Studies}
\label{subsec:ablation}
To validate the design choices of DR-Arena, we conduct ablation studies focusing on three core components: \rc{tree-guided task generation, rubric-based judgments, and the evolvement loop.}

\paragraph{\rc{Tree-guided Task Generation.}}

\begin{table}[h]
\centering
\small
\renewcommand{\arraystretch}{1.2}
\resizebox{\linewidth}{!}{
\begin{tabular}{l c c}
\toprule
\textbf{Configuration} & \textbf{Question Quality} & \textbf{Rubric Accuracy} \\
\midrule
\textbf{Full Tree (Ours)} & \textbf{89.0\%} & \textbf{88.0\%} \\
No Logic Chain & 2.0\% & 8.0\% \\
Flat Context & 9.0\% & 4.0\% \\
\bottomrule
\end{tabular}
}
\caption{{Human Preference Rates on Task Generation.}}
\label{tab:human_study_tree}
\vspace{-3mm}
\end{table}

To verify the \rc{effectiveness of Information Trees,}
we conduct a blind human study comparing tree-based \rc{task generation}
against two ablation baselines:
(1) Flat Context: Providing the Examiner with unstructured search snippets without topological edges.
(2) No Logic Chain: Providing node content but withholding ancestor context (the reasoning path). 
For 50 sampled cases, two expert annotators performed a blind three-way selection to choose the generated question that best necessitated both Multi-hop Logic (Depth) and Multi-source Synthesis (Width) and the verification checklist that best captured the ground truth. \rc{Results are presented in Table \ref{tab:human_study_tree}.}
\rc{We observe that the full tree approach achieves the best performance,} which \rc{has}
a selection rate of 89\% for question generation and 88\% for rubric construction. In contrast, the Flat Context baseline often degenerates into shallow factoid lookups, confirming that topological structure is essential for constructing valid deep research tasks.

\begin{table}[t]
\centering
\small
\renewcommand{\arraystretch}{1.2}
\resizebox{0.99\linewidth}{!}{
\begin{tabular}{l c c}
\toprule
\textbf{Configuration} & \textbf{Pearson} & \textbf{Spearman} \\
\midrule
\textbf{DR-Arena (Full)} & \textbf{0.74} & \textbf{0.94} \\
\midrule
\textit{Impact of Adjudication} & & \\
w/o Rubric (Intuition Judge) & 0.72 & 0.83 \\
\midrule
\textit{Impact of Evolvement Loop} & & \\
w/o Evolvement (Round 1 only) & 0.41 & 0.77 \\
w/o Evolvement (Round 2 only) & 0.68 & 0.94 \\
\bottomrule
\end{tabular}
}
\caption{\textbf{Ablation Results.} Removing the \rc{rubrics or the} evolvement loop
significantly degrades alignment.}
\label{tab:ablation}
\end{table}

\begin{figure}[t]
    \setlength{\abovecaptionskip}{3pt} 
    \setlength{\belowcaptionskip}{0pt}
\centering
\includegraphics[width=0.6\columnwidth]{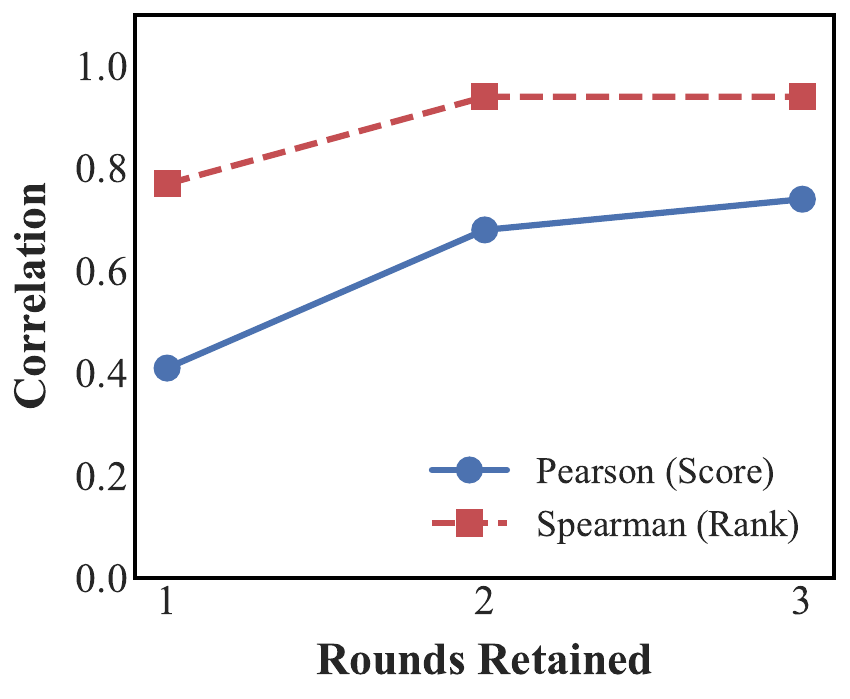}
\caption{{Evolution of Correlation across Rounds.}}
\label{fig:ablation_rounds}
\vspace{-3mm}
\end{figure}

\paragraph{Rubric-Based Judgments.}
We evaluate the adjudication mechanism by removing the generated Checklists and forcing the Examiner to judge based on internal knowledge and optional 
search. \rc{As shown in Table \ref{tab:ablation},} the Spearman correlation drops from 0.94 to 0.83. \rc{When examining the mismatched judgments closely, we see that,} while the intuition-based judge can capture stylistic quality, it struggles to detect subtle hallucinations in synthesis. This demonstrates that Evidence-Based Adjudication is critical for reducing judge hallucination and ensuring rigorous alignment with human standards.

\paragraph{The Evolvement Loop.}
We assess the impact of the multi-round dynamic investigation by \rc{cutting off investigation rounds early and present the results in Table \ref{tab:ablation}.}
The results derived solely from the First Round are insufficient\rc{, where Spearman correlation drops from 0.94 to 0.77}, confirming that single-turn interactions fail to distinguish closely matched models, often resulting in ambiguous ties or noisy verdicts. By Round 2, the Spearman ranking correlation jumps to 0.94, demonstrating that the pressure test mechanisms are highly efficient at sorting agents into the correct hierarchy.
\rc{The loop remains essential beyond this point, which we show with a more straightforward plot in Figure \ref{fig:ablation_rounds}:}
The Pearson correlation, which measures the linearity of the score gaps, continues to improve from 0.68 to 0.74. This confirms that the subsequent rounds are important for calibrating the magnitude of the performance difference. 

\paragraph{Cross-Examiner Validation.}
To further examine whether the reliability of DR-Arena depends on a specific Examiner model, we conduct a Cross-Examiner validation experiment. We randomly sample 50 matches from the tournament logs and re-adjudicate them using two alternative frontier LLMs, {GPT-5.2-Chat} and {Claude-Opus-4.6}. We then compare their verdicts against the original adjudications produced by Gemini-3-Pro.

As shown in Table \ref{tab:cross_examiner}, the alternative Examiners achieve high agreement with the original Examiner, with Percent Agreement rates of 93.02\% and 88.70\%, and Cohen's Kappa scores of 0.901 and 0.839, respectively. These results indicate that DR-Arena's judgments are highly consistent across diverse frontier LLMs, suggesting that the evaluation signal is not primarily driven by the stylistic bias of a particular judge model. We hypothesize that this robustness stems from the structured nature of our framework: unlike fully open-ended judging, the Information Tree provides grounded rubrics that function as an explicit answer key, substantially reducing variance in the adjudication process.

\begin{table}[t]
\centering
\small
\renewcommand{\arraystretch}{1.2}
\resizebox{0.99\linewidth}{!}{
\begin{tabular}{l c c}
\toprule
\textbf{Judge Model} & \textbf{Percent Agreement} & \textbf{Cohen's Kappa} \\
\midrule
GPT-5.2-Chat     & 93.02\% & 0.901 \\
Claude-Opus-4.6  & 88.70\% & 0.839 \\
\bottomrule
\end{tabular}
}
\caption{\textbf{Cross-Examiner Validation.} Agreement rates between alternative Examiners and the original Examiner on 50 sampled matches.}
\label{tab:cross_examiner}
\vspace{-3mm}
\end{table}

\subsection{Validation via Human Study}

\begin{table}[t]
\centering
\small
\renewcommand{\arraystretch}{1.2}
\resizebox{\columnwidth}{!}{
\begin{tabular}{l l c}
\toprule
\textbf{Component} & \textbf{Evaluation Criterion} & \textbf{Result} \\
\midrule
\multirow{2}{*}{\textbf{Task Generation}} 
 & Question Validity (Structure) & 90.6\% \\
 & Rubric Factuality (Ground Truth) & 89.1\% \\
\midrule
\multirow{3}{*}{\textbf{Examiner Process}} 
 & Verdict Alignment (Cohen's $\kappa$) & 0.91 \\
 & Transition Logic Accuracy & 96.9\% \\
 & Stop Condition Efficiency & 92.2\% \\
\midrule
\textbf{Human Annotators} & Inter-Annotator Agreement (Cohen's $\kappa$) & 0.88 \\
\bottomrule
\end{tabular}
}
\caption{\textbf{Human Audit Results.} Expert annotators ($N=2$) evaluated the quality of automated components across 30 randomly sampled matches (64 turns).}
\label{tab:human_audit_results}
\vspace{-3mm}
\end{table}

To validate the reliability of our automated components, we conduct a rigorous human study where two expert annotators audited a random sample of 30 full-match logs (covering 64 interaction turns). Detailed \rc{human study setups and interface demos}
are provided in Appendix \ref{app:human_study_details}.

\rc{During the study, we ask the annotators to evaluate the match quality while focusing on two aspects: task generation and Examiner judgment. The results are summarized in Table \ref{tab:human_audit_results}, which}
strongly validate \rc{the quality of} both the generation and adjudication pipelines.
First, regarding {Task Generation}, 90.6\% of the generated questions are confirmed to strictly adhere to the Depth and Width 
principles and 89.1\% of the rubrics are verified as factually correct based on the source URLs. 
Second, regarding the {judgment process}, the automated verdicts achieve substantial alignment with human experts\rc{' judgments}, yielding a Cohen's Kappa agreement of 0.91. The two human annotators themselves achieve a Cohen's Kappa agreement of 0.88 on verdict selection, indicating high inter-annotator consistency. Furthermore, the adaptive mechanism is proved to be highly reliable: 96.9\% of the Evolvement Loop transitions correctly follow the diagnostic logic, and 92.2\% of the matches are deemed to have stopped at an efficient round count. These findings 
validate the automated Examiner as a trustworthy, scalable proxy for human adjudication.

\section{In-depth Analysis}
\label{sec:analysis}

\rc{Beyond quantitative evaluation results, we also}
perform in-depth analysis to examine the efficiency of the dynamic loop and the error distribution profiles of different models. Extended statistical diagnostics, including the aggregate distribution of 
verdicts and the structural diversity of the generated Information Trees, are detailed in Appendix \ref{app:diagnostics}.

\subsection{Efficiency: The Skill Gap vs. Rounds}
\begin{figure}[t]
    \setlength{\abovecaptionskip}{3pt} 
    \setlength{\belowcaptionskip}{0pt}
\centering
\includegraphics[width=0.95\columnwidth]{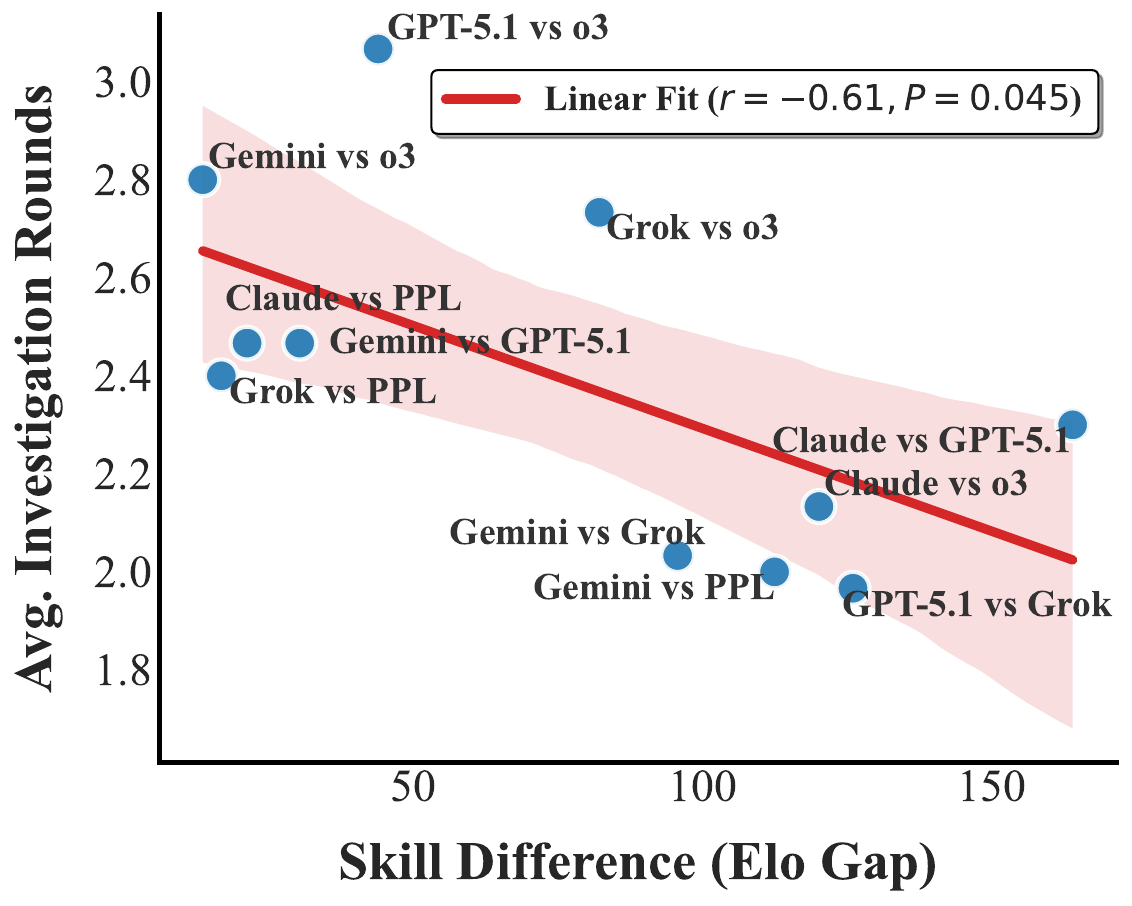} 
\caption{Relationship between Skill Gap and Rounds.}
\label{fig:efficiency_scatter}
\vspace{-3mm}
\end{figure}

To examine the hypothesis of DR-Arena where follow-up rounds make performance gaps more apparent, we analyze the relationship between \textit{skill difference} (Elo Gap) and \textit{number of investigation rounds} in Figure \ref{fig:efficiency_scatter}. 
We quantify the linear relationship using the Pearson correlation coefficient ($r$) and assess statistical significance via the $p$-value. We observe a significant negative correlation ($r = -0.61$, $p = 0.045$) \rc{between average investigation rounds and skill gap, validating that pairs with smaller skill gaps indeed require more investigation rounds}. \rc{In the graph, we see that} the system converges rapidly for high-divergence pairs; for instance, matches involving \textit{GPT-5.1-Search} {vs.} \textit{Grok-4-Search} ($\Delta \text{Elo} \approx 126$) average fewer than 2.0 rounds, as the stronger model quickly satisfies hard constraints while the weaker fails. Conversely, closely matched pairs such as \textit{Gemini-2.5-Pro} vs. \textit{o3-Search} ($\Delta \text{Elo} \approx 13.5$) extend to 
2.8 rounds. In these scenarios, the system automatically detects and triggers the Pressure Test mechanism, escalating complexity to force a fine-grained differentiation. \rc{Therefore,} DR-Arena effectively functions as an efficient sorting algorithm, concentrating computational cost 
on the decision boundaries where it yields the highest information gain.

\rc{Notable outliers provide further insights into model architectures. The matchup between \textit{GPT-5.1-Search} and \textit{o3-Search} exhibits the highest investigation duration (3.07 rounds) despite a moderate Elo gap ($\Delta \text{Elo} \approx 44$). We hypothesize this stems from Behavioral Homogeneity within the OpenAI family: Both models likely share similar data and reasoning answer patterns, which lead to more ties. The system then escalates difficulty with more rounds to uncover subtle differences in long-horizon synthesis capabilities, demonstrating that DR-Arena can distinguish even highly correlated models, albeit at higher computational cost.}

\subsection{Error Distribution: Depth vs. Width}
\label{subsec:error_dist}
\begin{figure}[t]
    \setlength{\abovecaptionskip}{3pt} 
    \setlength{\belowcaptionskip}{0pt}
    \centering
    \includegraphics[width=\linewidth]{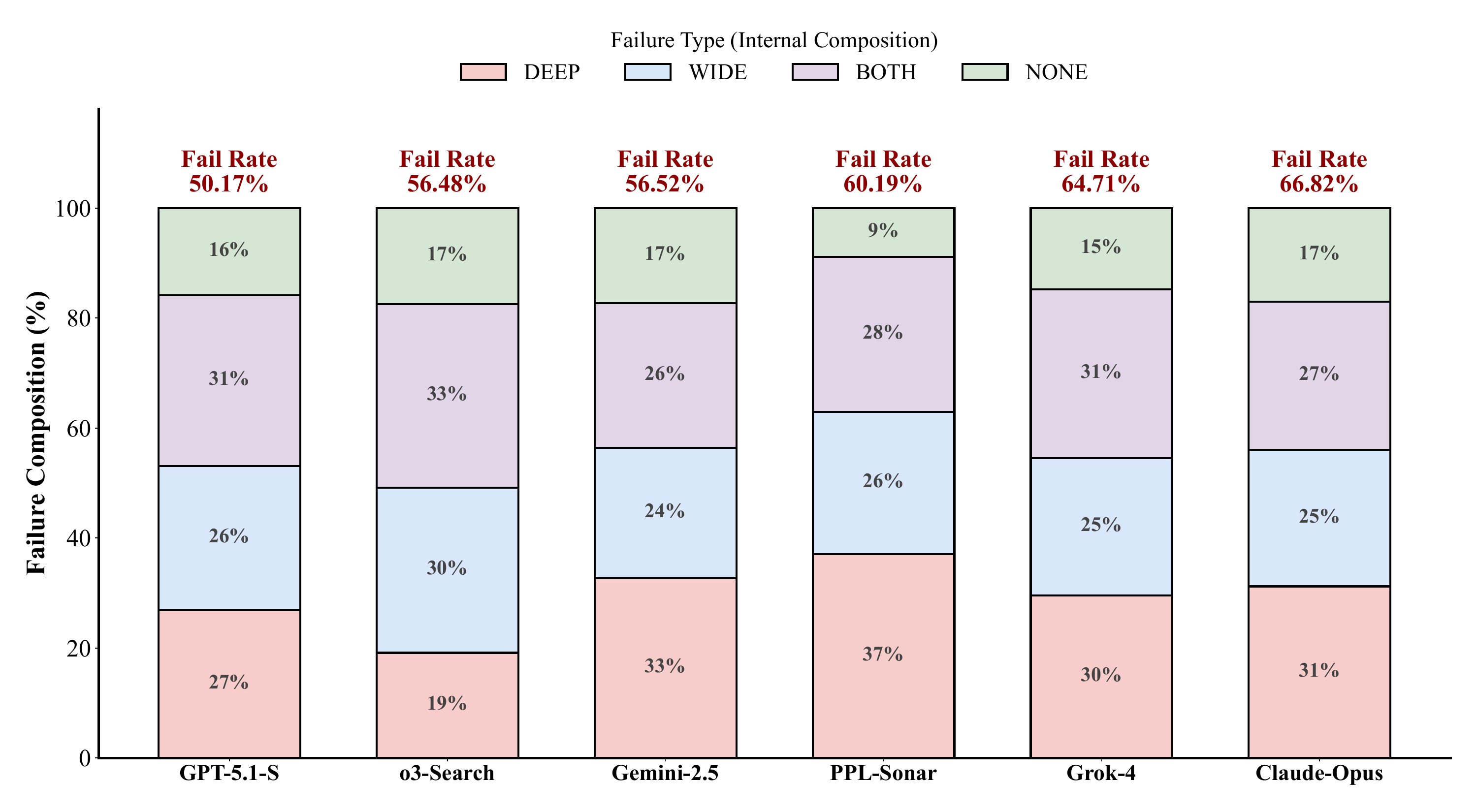}
    \caption{{Per-Model Performance Profile.} Models are ordered by their overall failure rates (at the top).}
    \label{fig:model_perf}
    \vspace{-3mm}
\end{figure}

\rc{To provide a closer look into each model's performance profile, we show a dissection of failure types in Figure \ref{fig:model_perf}.}
We define the Failure Rate as the ratio of non-winning rounds (including losses and ties) to the total number of rounds each model participated in. Ordering models by this metric reveals a clear robustness hierarchy, with \textit{GPT-5.1-Search} demonstrating the highest stability with a minimum failure rate of 50.17\%.

\rc{Beyond aggregate rates, the normalized failure distribution exposes architectural trade-offs. \textit{GPT-5.1-Search} exhibits symmetric errors (27\% Depth, 26\% Width), suggesting balanced capabilities without structural bottlenecks. In contrast, \textit{o3-Search} shows asymmetry favoring logical deduction: despite higher total failures, it has the lowest \textsc{Depth} failure rate (19\%), with limitations primarily in coverage breadth (\textsc{Width}: 30\%). Models like \textit{PPL-Sonar-Pro} and \textit{Grok-4} display the inverse pattern, with failures skewed toward \textsc{Deep} reasoning deficits (37\% and 30\% respectively), indicating effective retrieval but inconsistent logical reasoning. The persistent \textsc{Both} failures ($\sim$20-30\%) across all models underscore the difficulty of simultaneously satisfying depth and width constraints. These profiles reveal capability trade-offs beyond aggregate metrics, enabling targeted model improvements.}

\section{Conclusion}

\rc{We introduce DR-Arena, a fully automated framework for evaluating DR agents in dynamic environments. Existing static benchmarks suffer from temporal misalignment, penalizing agents that retrieve current information contradicting outdated ground truth. DR-Arena addresses this by constructing real-time Information Trees synchronized with the live internet. Beyond timeliness, it transforms evaluation into an active stress test through Adaptive Evolvement, efficiently isolating performance boundaries and disentangling failures in logical deduction (\textit{Depth}) from deficits in information aggregation (\textit{Width}). Achieving 0.94 Spearman correlation with human-verified leaderboards, DR-Arena offers a scalable, reliable alternative to costly manual evaluation for benchmarking next-generation autonomous research systems.}



\section*{Limitations}

While DR-Arena provides a scalable alternative to human annotation, we acknowledge several limitations inherent to the framework.

First, although our cross-examiner validation shows strong agreement across different judge models, DR-Arena still relies on a single Examiner in the main evaluation pipeline, so judge-specific preferences cannot be fully ruled out.

Second, when the generated ground-truth rubrics conflict with the Examiner's internal parametric knowledge, there arises a risk that a judge could override the provided rubrics due to strong parametric priors. In some cases, the judge could also successfully disregard a mistake in the generated list, as shown in {Appendix \ref{app:case_study_beneficial}}. Overall, the judge's parametric knowledge conflicts remain a risk and future direction to explore.

Third, because DR-Arena operates over the live web and depends on commercial search APIs, exact reproduction may be affected by temporal changes in indexing, ranking, geoblocking, and webpage availability. This creates a trade-off between ecological realism and fully controlled comparison: while we share the same fixed Information Trees across agents for fair evaluation, source prioritization and search ranking may still disadvantage agents that rely on different but potentially more reliable sources. We reduce this risk by biasing tree construction toward high-reliability seed domains, while noting that fully freezing the search environment would re-introduce contamination risks. 

Finally, although our evidence-based rubric correlates well with human preference on fact-heavy research tasks, it may undervalue creative synthesis or lateral thinking that deviates from the strict logical path of the generated tree.

\section*{Acknowledgments}
This research/project is supported by the National Research Foundation Singapore under the AI Singapore Programme (AISG Award No: AISG3-RPGV-2025-016). This research/project is supported by the National Research Foundation, Singapore under its AI Singapore Programme (AISG Award No: AISG2-PhD-2021-01-001).

\bibliography{custom}
\clearpage

\appendix

\section{Prompts and Configuration Settings}
\label{app:prompts}
To ensure reproducibility, we provide all prompt templates used in DR-Arena, including Automated Task Generation and Evidence-Based Judgment. In the templates below, dynamic content injected by the system at runtime is denoted by \placeholder{Blue Placeholders}.

\subsection{Automated Task Generation}
The Examiner uses the following prompt to transform a raw Information Tree into a structured ``Depth \& Width'' research task. The system dynamically calculates word count constraints based on the tree depth and width.

\begin{promptbox}{System Prompt: Task Generation}
# TASK: Generate a ``Depth & Width'' Search Evaluation Query

You are an expert at creating complex, multi-hop search queries designed to test the limits of Search Agents. Your goal is to synthesize a question that requires |\textbf{Logical Reasoning (Depth)}| to identify the subjects and |\textbf{Broad Information Aggregation (Width)}| to answer fully.

--- 1. THE HIDDEN KNOWLEDGE (Source Material) ---
*Note: This content is hidden from the test taker. It is only for you to formulate the question and the grading criteria.*

|\textbf{OVERALL DOMAIN/TOPIC}|: ``|\placeholder{Root Topic}|''
|\textbf{A. Reasoning Chain (Background/Context)}|:
|\placeholder{Serialized Reasoning Nodes (Ancestors)}|
|\textbf{B. Target Answers (The Facts to Retrieve)}|:
|\placeholder{Serialized Target Nodes (Siblings)}|

--- 2. QUESTION GENERATION STEPS (READ CAREFULLY) ---
|\textbf{Rule 1: ABSOLUTE GROUNDING - CRITICAL}|
- |\textbf{YOU MUST}| generate the question based |\textbf{ONLY}| on the specific entities and facts found in the [Hidden Knowledge] above.
- |\textbf{STRICT PROHIBITION:}| Do NOT ignore the provided text.
- |\textbf{Relevance}|: The question MUST be relevant to the |\textbf{Overall Domain/Topic}| (''|\placeholder{Root Topic}|''). Do not hallucinate unrelated topics.

|\textbf{Rule 2: COMPLETE DE-CONTEXTUALIZATION (No Leaking)}|
- |\textbf{FORBIDDEN:}| You MUST NOT mention the specific filename, website title, directory name, or document header found in the source.
- |\textbf{REQUIRED:}| Treat the provided text as just *one instance* of a universal fact. Ask about the *entities themselves*, not about the *document* describing them. 
- |\textbf{Litmus Test:}| If the user needs the specific JSON file you read to understand the question, YOU FAILED. The question must be solvable using Google/Bing to find the *original primary sources*.

|\textbf{STEP 1: Deep Reasoning (The Filter)}|
- Analyze the [Reasoning Chain] to identify the specific logic, condition, or category that groups the target entities together.
- |\textbf{RULE}|: Do NOT mention the specific names of the [Target Entities] in the question.
- |\textbf{RULE}|: Use the [Reasoning Chain] logic to strictly define the group.

|\textbf{STEP 2: Wide Aggregation (The Scope)}|
- If the [Target Answers] contain multiple entities, the question MUST require reading and comparing information from |\textbf{ALL}| of them.
- The answer must not be resolvable by finding a single document; it must require aggregating details across all identified targets.

|\textbf{STEP 3: Synthesis (The Depth \& Width Question)}|
- Combine Step 1 and Step 2 into a single, cohesive natural language question.
- |\textbf{CRITICAL}|: Ensure the question targets |\textbf{Publicly Verifiable Facts}|. Do not ask about obscure details that exist *only* within the specific phrasing of the provided source text. The question must be answerable by searching external, general web sources.

--- 3. CHECKLIST DEFINITIONS (CRITICAL) ---
|\textbf{STEP 1 Draft the Gold Standard Answer}|: Formulate a complete answer based on the [Hidden Knowledge].
|\textbf{STEP 2 Extract Checklists}|: Deconstruct the answer into specific verification points.
    - |\textbf{Checklist Width (Completeness \& Details)}|: |\textbf{Content}|: The Specific Attributes/Facts requested in the query. |\textbf{Purpose}|: Once the entity is found, did the agent gather *all* the requested scattered details?
    - |\textbf{Checklist Depth (Identity \& Logic)}|: |\textbf{Content}|: The Correct Entity Names + The Logic Validation. |\textbf{Purpose}|: Did the agent use the reasoning chain to find the *correct* person/thing?

--- 4. OUTPUT FORMAT (JSON) ---
Return the result in the following JSON format:
{
    ``question'': ``The final Depth & Width search query'',
    ``word_limit_instruction'': ``|\placeholder{Dynamic Constraint String}|'',
    ``checklist_width'': [
        ``Specific Detail A for Entity 1'',
        ``Specific Detail B for Entity 1'',
        ...
    ],
    ``checklist_depth'': [
        ``Target Entity 1 Name + Logic Proof'',
        ...
    ],
    ``rationale'': ``Briefly explain how the question uses logic to mask entities (Depth) and requests scattered info (Width).''
}
\end{promptbox}

\subsection{Evidence-Based Judgment}
The Examiner acts as a Judge to evaluate two DR agents responses using the generated Rubrics.

\begin{promptbox}{System Prompt: Judgment}
### Role: Super-User Evaluator (Simulating Human Preference)
Compare Response A and Response B to identify which search agent provides a better USER EXPERIENCE.
While accuracy is paramount, you must also heavily weigh |\textbf{comprehensiveness, formatting, and helpfulness}| -- traits that human users value in search engines like Perplexity, Gemini, or SearchGPT.

--- 1. QUERY & CONSTRAINT ---
Query: |\placeholder{Question}|
Constraint: |\textbf{Maximum \placeholder{Word Limit} words}|. (Note: Do not penalize slightly going over if the quality is high. Only penalize extreme verbosity).

--- 2. GROUND TRUTH CHECKLIST ---
[WIDTH-Completeness]: |\placeholder{Checklist Width JSON}|
[DEPTH-Logic]: |\placeholder{Checklist Depth JSON}|

--- 3. RESPONSES ---
=== Agent A ===
(Citation Count: |\placeholder{Count A}|)
|\placeholder{Final Answer A}|
=== Agent B ===
(Citation Count: |\placeholder{Count B}|)
|\placeholder{Final Answer B}|

--- 4. EVALUATION CRITERIA (Aligned with Human Preference) ---
|\textbf{Dimension 1: Accuracy (The Foundation)}|
- |\textbf{Core Entity Check}|: Determine if each agent passes the DEPTH Logic (Found the right entity?). (If BOTH fail this, it's a LOW TIE).
- |\textbf{Sub-Point Accuracy}|: Did the agent answer *all* parts of the prompt correctly? Determine if each agent passes the WIDTH Aggregation (Found the specific details?).
- If BOTH agents have significant hallucinations (even on different parts), consider a |\textbf{Low Quality Tie}|.

|\textbf{Dimension 2: User Utility \& Completeness (The Experience)}|
- |\textbf{Helpfulness}|: Is the answer easy to read? Does it actually solve the user's underlying intent?
- |\textbf{Information Density}|: Unlike simple chatbots, Search Agents should provide |\textbf{rich context}|.
- |\textbf{Helpful Recovery}|: If the exact answer isn't in the context, did the agent try to synthesize *related* useful info?
- |\textbf{Citation Density}|: A higher citation count is generally preferred as it indicates better groundedness.

|\textbf{Dimension 3: Presentation \& Structure }|
- |\textbf{Markdown Mastery}|: REWARD the use of |\textbf{Bold}| headers, Bullet points, and Tables.
- |\textbf{Scannability}| \& |\textbf{Directness}|: Can a user find the specific answer in 2 seconds? (BLUF - Bottom Line Up Front)?

--- 4. SCORING RUBRIC ---
- |\textbf{[[A/B\_MUCH\_BETTER]] (+2)}|:
    - The winner found the correct Entity AND answered sub-points correctly (No Hallucinations).
    - The loser failed the Depth Logic (Wrong Entity) or missed major Checklists.
    - *Note: Do not give MUCH_BETTER if the winner has a factual error in a sub-point.*
- |\textbf{[[A/B\_BETTER]] (+1)}|:
    If winner has errors, cap at BETTER.
    - |\textbf{The ``Flawed Winner''}|: The winner got the Main Entity right, but missed a detail or hallucinated on a minor sub-point. The loser failed the Main Entity.
    - |\textbf{The ``Style Winner''}|: Both are factually accurate, but one has significantly better formatting/comprehensiveness.
    - |\textbf{The ``Nuance Winner''}|: Both failed slightly, but the winner's failure was less catastrophic than the loser's.
- |\textbf{[[Tie]]}|:
    - *High Quality*: Both gave perfect, well-formatted, accurate answers.
    - *Low Quality*: Both failed to find the core entity or both hallucinated significantly.

|\textbf{Error Diagnosis}|
- If there is a loser, identify WHY they lost.
- |\textbf{DEPTH}|: Failed logic/identity (Wrong Entity).
- |\textbf{WIDTH}|: Failed detail aggregation (Missing Facts).
- |\textbf{BOTH}|: Failed both depth logic and width details.
- |\textbf{NONE}|: No hard checklist failures, when the winner won solely on Soft Filters like citations/formatting.

--- 5. OUTPUT FORMAT (JSON) ---
{
    ``verdict'': ``[[A_MUCH_BETTER]]'' OR ``[[A_BETTER]]'' OR ``[[Tie]]'' ... ,
    ``tie_quality'': ``HIGH'' OR ``LOW'' OR ``N/A'',
    ``loser_failure_type'': ``DEPTH'' OR ``WIDTH'' OR ``BOTH'' OR ``NONE'',
    ``reasoning'': ``First, verify Depth Logic for both. Then, compare Width/Completeness...''
}
\end{promptbox}

\newpage
\section{Example Generated Tasks}
\label{app:example_tasks}

This section presents five diverse examples of ``Depth \& Width'' research tasks generated by the Examiner. These examples demonstrate the system's ability to construct complex queries requiring multi-hop reasoning (Depth) and broad information aggregation (Width) across different domains.

\begin{promptboxtext}{Example 1: Local Business Structure Analysis}
\textbf{[QUESTION]} \\
Locate the gymnastics organization in the New York/New Jersey area that structures its competitive teams into three specific tiers: an in-house 'Club Team' that competes exclusively against the organization's other branches, a 'USA-IGC' program capped at 3 training days per week to accommodate other sports, and a third high-performance program that explicitly prohibits athletes from participating in other activities. Provide the name of this organization, the specific name of the high-performance program, and a list of all the cities where they currently operate gyms.

\vspace{0.5em}
\hrule
\vspace{0.5em}

\textbf{[CHECKLIST / GRADING KEY]} \\
\textbf{Entities to find}: Target Entity: \textbf{Gold Medal Gymnastics} (or GMGC / Gold Medal Gymnastics \& Ninja). \\
\textbf{Logic Proof}: Identified via the unique combination of an internal-only 'Club Team' and a 'Junior Olympic' program that prohibits other sports.

\textbf{Specific Details}:
\begin{itemize}[leftmargin=*, nosep]
    \item High-Performance Program Name: \textbf{Junior Olympic Program} (or Junior Olympic Team)
    \item City 1: Centereach
    \item City 2: Garden City
    \item City 3: Huntington
    \item City 4: Levittown
    \item City 5: Rocky Point
    \item City 6: Short Hills
    \item City 7: Smithtown
\end{itemize}
\end{promptboxtext}

\begin{promptboxtext}{Example 2: Cross-Domain Synthesis (Marketing \& Labor Stats)}
\textbf{[QUESTION]} \\
Locate the analysis by 'Marketing Eye Atlanta' regarding a viral charity campaign that involved a 'bucket of iced water' and the participation of 'the most feared woman in fashion.' Identify the campaign and list the five specific elements cited that made it go viral. Then, identifying the specific occupation defined by the Bureau of Labor Statistics (BLS) as professionals who 'create and maintain a positive public image for the clients they represent,' provide the following data points from the 2024 occupational profile: the 2024 median annual pay, the job outlook percentage for the 2024–34 decade, and the projected numeric employment change for that same period.

\vspace{0.5em}
\hrule
\vspace{0.5em}

\textbf{[CHECKLIST / GRADING KEY]} \\
\textbf{Entities to find}: Target Entity 1: \textbf{ALS Ice Bucket Challenge} (identified via Marketing Eye Atlanta \& Anna Wintour clue), Target Entity 2: \textbf{Public Relations Specialists} (identified via job description match).

\textbf{Specific Details}:
\begin{itemize}[leftmargin=*, nosep]
    \item Element 1: Emotional AND logical appeal
    \item Element 2: Celebrity certification
    \item Element 3: Amusement
    \item Element 4: Simplicity
    \item Element 5: Shareability
    \item 2024 Median Annual Pay: \textbf{\$69,780}
    \item Job Outlook (2024–34): \textbf{5\%} (Faster than average)
    \item Employment Change (2024–34): \textbf{15,000}
\end{itemize}
\end{promptboxtext}

\begin{promptboxtext}{Example 3: Tech Policy \& International Specifications}
\textbf{[QUESTION]} \\
Analyze the language support specifications for the 'Live Translation' feature powered by Apple Intelligence as detailed in the user guide covering macOS Tahoe and iOS 26. Determine which three languages are explicitly compatible with Live Translation in the Messages app but are not listed as supported for Live Translation in Phone and FaceTime calls. Following this, consult the terms and conditions for the Apple Store Singapore to find the specific return policy for products purchased in volume (defined as an aggregate of more than four items). Provide the names of the three distinct languages, the number of days allowed for a volume return, and the restocking fee percentage charged.

\vspace{0.5em}
\hrule
\vspace{0.5em}

\textbf{[CHECKLIST / GRADING KEY]} \\
\textbf{Entities to find}: Target Entity: \textbf{Apple Intelligence Live Translation} (Logic: Compare Messages support list vs. Phone/FaceTime support list), Target Entity: \textbf{Apple Store Singapore Policy} (Logic: Identify specific terms for 'returned in volume' > 4 items).

\textbf{Specific Details}:
\begin{itemize}[leftmargin=*, nosep]
    \item Language 1: \textbf{Dutch}
    \item Language 2: \textbf{Turkish}
    \item Language 3: \textbf{Vietnamese}
    \item Volume Return Deadline: \textbf{7 days}
    \item Volume Restocking Fee: \textbf{25\%}
\end{itemize}
\end{promptboxtext}

\begin{promptboxtext}{Example 4: Medical Service Scope \& Pricing}
\textbf{[QUESTION]} \\
Investigate the remote medical second opinion service operated by the joint venture between Cleveland Clinic and Amwell. Provide a detailed breakdown of its geographic scope by listing the U.S. states eligible for the full 'Concierge Plus' virtual visit, those restricted to the written report only, and the specific states where the service is unavailable. Additionally, report the service costs for U.S. patients versus international patients, and enumerate the specific countries where the international service is prohibited.

\vspace{0.5em}
\hrule
\vspace{0.5em}

\textbf{[CHECKLIST / GRADING KEY]} \\
\textbf{Entities to find}: Target Entity: \textbf{Virtual Second Opinions (VSO)} by 'The Clinic' (Cleveland Clinic \& Amwell Joint Venture).

\textbf{Specific Details}:
\begin{itemize}[leftmargin=*, nosep]
    \item Identifies 'Concierge Plus' (Virtual Visit) states: Ariz., Calif., Colo., Conn., Fla., Ga., Ill., Ind., Ky., Mich., N.C., N.J., N.Y., Ohio, Pa., S.C., Tenn., Texas, Va., Wisc., W.V.
    \item Identifies 'Written Report Only' states: Alaska, Ala., Ark., D.C., Del., Hawaii, Iowa, Idaho, Kan., La., Mass., Md., Minn., Mo., Miss., Mont., N.D., Neb., N.H., N.M., Nev., Okla., Ore., Utah, Vt., Wash., Wyo.
    \item Identifies Excluded U.S. states: \textbf{Maine, Rhode Island (R.I.), South Dakota (S.D.)}
    \item States U.S. Pricing: \textbf{\$1,690} (Report only) and \textbf{\$1,990} (Report + Virtual Visit)
    \item States International Pricing: \textbf{\$4,500 USD}
    \item Lists Excluded Countries: Australia, China, Germany, Denmark, Greece, Iran, North Korea, South Korea, Kazakhstan, Malaysia, Russian Federation, Sweden, Turkey
\end{itemize}
\end{promptboxtext}

\begin{promptboxtext}{Example 5: Cultural Analysis (Music History)}
\textbf{[QUESTION]} \\
In a 1988 review by Steve Pond (RS 537), two distinct Los Angeles-bred acts were compared: one described as an ``unprolific'' artist making ``immaculate pop music'' with lushness borrowed from soundtracks, and the other a ``young and restless'' band labeled the ``true heir to Led Zeppelin'' but stripped of ``fairy-tale whimsy.'' Despite their differences, the reviewer noted that both artists' respective albums from that year were populated by ``recognizable, real people.'' Identify these two acts and their corresponding albums. Then, based on the specific descriptions in the review, list the following tracks: for the band, the two songs characterized as ``hard-boiled riff rockers'' and the acoustic song deemed a ``worthy Left Coast successor to Walk on the Wild Side''; for the songwriter, the two songs presenting ``bucolic views of a childhood in New Orleans,'' the two ``naive, devoted love songs,'' and the final track described as a ``chilling, coldblooded moment'' involving a message to his son.

\vspace{0.5em}
\hrule
\vspace{0.5em}

\textbf{[CHECKLIST / GRADING KEY]} \\
\textbf{Entities to find}: Band: \textbf{Jane's Addiction} (Album: Nothing's Shocking), Songwriter: \textbf{Randy Newman} (Album: Land of Dreams).
\textbf{Logic Proof}: Validates the ``heir to Led Zeppelin'' vs ``immaculate pop'' comparison from the 1988 Steve Pond review.

\textbf{Specific Details}:
\begin{itemize}[leftmargin=*, nosep]
    \item Band Song (Hard-boiled): ``Had a Dad''
    \item Band Song (Hard-boiled): ``Standing in the Shower... Thinking''
    \item Band Song (Successor): ``Jane Says''
    \item Songwriter Song (Bucolic): ``Dixie Flyer''
    \item Songwriter Song (Bucolic): ``New Orleans Wins the War''
    \item Songwriter Song (Love): ``Falling in Love''
    \item Songwriter Song (Love): ``Something Special''
    \item Songwriter Song (Chilling): ``I Want You to Hurt Like I Do''
\end{itemize}
\end{promptboxtext}

\newpage
\section{Algorithm of Evolvement Loop}
\label{app:algorithm}

The core logic of the DR-Arena Dynamic Investigation is detailed in Algorithm \ref{alg:evolvement}. The system manages the state of the Information Tree $T$, the current path $P$ (representing Depth), and the width constraint $W$.

\begin{algorithm}[H]
\small 
\caption{DR-Arena Adaptive Evolvement Loop}
\label{alg:evolvement}
\begin{algorithmic}[1]
\Require Information Tree $T$, Agents $M_A, M_B$, Examiner $E$
\Ensure Final Verdict $V$ and Scores $S_A, S_B$

\State \textbf{Initialize:} Path $P \leftarrow \text{RandomStart}(T)$, Width $W \leftarrow 2$
\State \textbf{Initialize:} Scores $S_A \leftarrow 0, S_B \leftarrow 0$

\While{$|S_A - S_B| < \text{Threshold}$ \textbf{and} $\text{Rounds} < \text{Max}$}
    
    \Statex \textcolor{gray}{\textit{// Phase 1: Environment Check \& Expansion}}
    \If{$|\text{Siblings}(P)| < W$}
        \State $T \leftarrow \text{Crawler.ExpandWidth}(P, \text{target}=W)$
    \EndIf
    
    \Statex \textcolor{gray}{\textit{// Phase 2: Context Extraction \& Task Generation}}
    \State $C_{depth} \leftarrow \text{Ancestors}(P)$
    \State $C_{width} \leftarrow \text{Siblings}(P, \text{limit}=W)$
    \State $Q, \mathcal{R} \leftarrow E.\text{Generate}(C_{depth}, C_{width})$ \Comment{Gen. Question \& Rubric}
    
    \Statex \textcolor{gray}{\textit{// Phase 3: Agent Execution \& Adjudication}}
    \State $Traj_A \leftarrow M_A(Q)$; \quad $Traj_B \leftarrow M_B(Q)$
    \State $V, F_{type} \leftarrow E.\text{Judge}(Traj_A, Traj_B, \mathcal{R})$
    \State Update $S_A, S_B$ based on $V$
    
    \Statex \textcolor{gray}{\textit{// Phase 4: Adaptive State Transition}}
    \If{$V$ is \textsc{Tie}}
        \If{$V$ is \textsc{Low\_Quality}} \Comment{Both Hallucinated}
            \State $P \leftarrow \text{Parent}(P)$ \Comment{\textbf{Backtrack}}
            \State $W \leftarrow \max(2, W - 1)$
        \Else \Comment{High Quality Tie}
            \State $W \leftarrow W + 1$ \Comment{\textbf{Pressure Test}}
            \State \Call{AttemptDescend}{}
        \EndIf
    \Else \Comment{Winner Decided}
        \If{$F_{type}$ is \textsc{Depth}} \Comment{Logic Failure}
            \State \Call{AttemptDescend}{}
        \ElsIf{$F_{type}$ is \textsc{Width}} \Comment{Coverage Failure}
            \State $W \leftarrow W + 1$
        \Else \Comment{Ambiguous Failure}
            \State $W \leftarrow W + 1$; \quad \Call{AttemptDescend}{}
        \EndIf
    \EndIf
\EndWhile

\Statex \hrulefill
\Procedure{AttemptDescend}{}
    \If{$P$ is Leaf Node}
        \State $T \leftarrow \text{Crawler.ExpandDepth}(P)$
    \EndIf
    \If{$P$ has Children}
        \State $P \leftarrow P + \text{RandomChild}(P)$
    \EndIf
\EndProcedure

\end{algorithmic}
\end{algorithm}

\newpage
\section{Full Match Walkthrough}
\label{app:full_match}

To illustrate the dynamic nature of the DR-Arena, we present a complete 5-round match trace between {Agent A (Perplexity-Sonar-Pro)} and {Agent B (Claude-Opus-4.1)}. 

This match, centered on the topic of ``Handheld Game Consoles,'' demonstrates the \textit{Adaptive Evolvement} mechanism: as agents succeed, the system increases complexity (Depth/Width); when they fail, the system targets their specific weaknesses. The match concludes via the \textit{Mercy Rule} when Agent B establishes a decisive lead.

\begin{promptboxtext}{Match Trace: Handheld Game Consoles (Rounds 1-5)}
\textbf{Match Metadata} \\
\textbf{Topic}: Handheld game console \\
\textbf{Agents}: Agent A (Perplexity-Sonar-Pro) vs. Agent B (Claude-Opus-4.1-Search) \\
\textbf{Final Result}: Agent B Wins (Score 5.0 vs 2.0)

\vspace{0.5em}
\hrule
\vspace{0.5em}

\textbf{=== ROUND 1 (Initialization) ===} \\
\textbf{State}: Depth 2 | Width 2 \\
\textbf{Question}: Identify the two handheld game consoles released in the 1990s that were uniquely designed to play the exact same physical game media as their manufacturer's home console counterparts. Provide launch price, battery count, and battery life.

\vspace{0.3em}
\textbf{Agent A Response}: Correctly identified \textbf{NEC TurboExpress} and \textbf{Sega Nomad}. Used bullet points for Price (\$249.99/\$179.99), Batteries (6 AA), and Life (~3 hrs). \\
\textbf{Agent B Response}: Correctly identified the same entities and data, but used narrative paragraphs.

\textbf{Verdict}: \textcolor{blue}{[[A\_BETTER]]} \\
\textbf{Reasoning}: ``Agent A is the winner based on Dimension 3 (Presentation). By using bullet points... Agent A made the answer immediately scannable... Agent B provided the same info but buried it within narrative paragraphs.''

\textbf{Evolution}: Winner is A $\rightarrow$ Action: \textbf{Pressure Test (Depth+1 \& Width+1)}

\vspace{0.5em}
\hrule
\vspace{0.5em}

\textbf{=== ROUND 2 (Complexity Increase) ===} \\
\textbf{State}: Depth 2 | Width 3 \\
\textbf{Question}: Identify the 8-bit handheld (1989) that succeeded 'Game \& Watch' and included \textit{Tetris}. Contrast battery specs against Atari Lynx and Sega Game Gear.

\vspace{0.3em}
\textbf{Agent A Response}: Identified Game Boy. Estimated battery life as ``10-15 hours'' based on early specs. Used placeholders for citations. \\
\textbf{Agent B Response}: Identified Game Boy. Estimated battery life as ``15-35 hours'' (closer to retrospective consensus). Provided valid URL citations.

\textbf{Verdict}: \textcolor{teal}{[[B\_BETTER]]} \\
\textbf{Reasoning}: ``Agent B wins based on Content Accuracy. The prompt asked for 'retrospective comparisons'... Agent B correctly captures the legendary efficiency (30+ hours) vs competitors... Agent A provided a conservative estimate.''

\textbf{Evolution}: Winner is B $\rightarrow$ Loser Failure: WIDE $\rightarrow$ Action: \textbf{Wide+1 (Increase Context Width)}

\vspace{0.5em}
\hrule
\vspace{0.5em}

\textbf{=== ROUND 3 (Technical Detail Test) ===} \\
\textbf{State}: Depth 2 | Width 4 \\
\textbf{Question}: Identify three commercial color handhelds (1989-1991). Provide Price, CPU Architecture, and Battery Count.

\vspace{0.3em}
\textbf{Agent A Response}: Identified Lynx, Game Gear, TurboExpress. Correctly identified Lynx CPU as \textbf{8-bit (6502 derivative)}. Excellent formatting. \\
\textbf{Agent B Response}: Identified same units. Incorrectly stated Lynx CPU was \textbf{``16-bit 65SC02''}.

\textbf{Verdict}: \textcolor{blue}{[[A\_BETTER]]} \\
\textbf{Reasoning}: ``Agent A wins on Technical Accuracy... Agent A correctly identified the Atari Lynx CPU as an 8-bit 6502 derivative. Agent B incorrectly stated it was a 16-bit processor (a common misconception).''

\textbf{Evolution}: Winner is A $\rightarrow$ Action: \textbf{Pressure Test}

\vspace{0.5em}
\hrule
\vspace{0.5em}

\textbf{=== ROUND 4 (Differentiation Test) ===} \\
\textbf{State}: Depth 2 | Width 5 \\
\textbf{Question}: Identify the Game Boy's two primary competitors (one 1989, one 1991).

\vspace{0.3em}
\textbf{Agent A Response}: Identified Competitors as \textbf{Atari Lynx} and \textbf{NEC TurboExpress}. \\
\textbf{Agent B Response}: Identified Competitors as \textbf{Atari Lynx} and \textbf{Sega Game Gear}.

\textbf{Verdict}: \textcolor{teal}{[[B\_MUCH\_BETTER]]} \\
\textbf{Reasoning}: ``Agent B correctly identified the Sega Game Gear as the primary competitor released in 1991... Agent A selected the NEC TurboExpress, which was a niche product and not a 'primary' competitor comparable to the Game Gear's market share.''

\textbf{Evolution}: Winner is B $\rightarrow$ Loser Failure: WIDE $\rightarrow$ Action: \textbf{Wide+1}

\vspace{0.5em}
\hrule
\vspace{0.5em}

\textbf{=== ROUND 5 (The Final Blow) ===} \\
\textbf{State}: Depth 2 | Width 6 \\
\textbf{Question}: Identify the dominant 8-bit console and its two main competitors (1989 \& 1991). Compare battery requirements. Describe the 1996 'Pocket' revision changes.

\vspace{0.3em}
\textbf{Agent A Response}: Again identified \textbf{NEC TurboExpress} as the main 1991 competitor. \\
\textbf{Agent B Response}: Correctly identified \textbf{Sega Game Gear}. Provided detailed comparisons of battery life (Game Boy 30h vs Game Gear 3-5h). Correctly described Pocket revision (AAA batteries, true black \& white screen).

\textbf{Verdict}: \textcolor{teal}{[[B\_MUCH\_BETTER]]} \\
\textbf{Reasoning}: ``Agent B correctly identified the 'Sega Game Gear'... Agent A incorrectly identified the 'NEC TurboExpress'. While the TurboExpress was released in 1991... it was not considered a 'main competitor'... The failure to identify the correct core entity is a critical logic failure.''

\vspace{0.5em}
\hrule
\vspace{0.5em}

\textbf{--- FINAL RESULT ---} \\
\textbf{Score}: Agent A (2.0) vs. Agent B (5.0) \\
\textbf{Status}: \textbf{[GAME OVER] Mercy Rule Triggered (Diff $\ge$ 2.0)} \\
\textbf{Winner}: \textbf{Agent B (Claude-Opus-4.1-Search)}
\end{promptboxtext}

\newpage
\section{Benchmarks}
\label{app:benchmarks}

\rc{We compare DR-Arena against six recent benchmarks for evaluating DR search agents. \textbf{BrowseComp} \citep{deng2024browsecomp} evaluates browsing and comprehension capabilities through curated web navigation tasks. \textbf{DeepResearch Bench} \citep{du2025deepresearchbench} assesses multi-step research abilities with 100 PhD-level research tasks. \textbf{LiveNewsBench} \citep{li2026livenews} and \textbf{LiveSearchBench} \citep{zhou2025livesearch} focus on real-time information retrieval from news and search results respectively. \textbf{LiveResearchBench} \citep{wang2025liveresearch} evaluates citation-grounded long-form reports with 100 expert-curated tasks. Finally, \textbf{Deep Research Bench} \citep{bosse2025DRsearch} evaluates web search capabilities with 89 multi-step web research tasks and a ``RetroSearch'' environment with a large frozen set of scraped web pages.}

\section{Dataset Statistics}
\label{app:dataset_stats}

We utilized a total of 30 Dynamic Information Trees constructed from Google Trends. The detailed distribution of these trees across different topics and web domains is summarized below.
\begin{table}[h]
    \centering
    \small
    \renewcommand{\arraystretch}{1.1}
    \caption{\textbf{Distribution of Source Domains.} The system successfully extracted verifiable structures from various authoritative sources, with Wikipedia serving as a major hub.}
    \label{tab:domain_dist}
    \begin{tabular}{l c}
        \toprule
        \textbf{Source Domain} & \textbf{Count} \\
        \midrule
        en.wikipedia.org & 8 \\
        gamefaqs.gamespot.com & 2 \\
        my.clevelandclinic.org & 2 \\
        support.apple.com & 1 \\
        www.creativebloq.com & 1 \\
        www.nyfa.edu & 1 \\
        www.kff.org & 1 \\
        gmgc.com & 1 \\
        www.finduslawyers.org & 1 \\
        demographicestimation.iussp.org & 1 \\
        www.nimh.nih.gov & 1 \\
        www.bls.gov & 1 \\
        info-ee.surrey.ac.uk & 1 \\
        www.laterpress.com & 1 \\
        www.becomeaprocomposer.com & 1 \\
        www.hhhistory.com & 1 \\
        online.nwmissouri.edu & 1 \\
        everettcc.libguides.com & 1 \\
        calpoison.org & 1 \\
        www.uscourts.gov & 1 \\
        open.umn.edu & 1 \\
        \bottomrule
    \end{tabular}
\end{table}
\begin{table*}[t]
    \centering
    \small
    \renewcommand{\arraystretch}{1.2}
    \caption{\textbf{Distribution of Information Trees by Topic.} The dataset covers a diverse range of high-level categories and specific sub-niches.}
    \label{tab:topic_dist}
    \begin{tabularx}{\textwidth}{X c}
        \toprule
        \textbf{Topic Hierarchy} & \textbf{Tree Count} \\
        \midrule
        Computers \& Electronics $>$ Software $>$ Operating Systems & 2 \\
        Games $>$ Computer \& Video Games $>$ Gaming Media \& Reference & 2 \\
        Law \& Government $>$ Legal & 2 \\
        Arts \& Entertainment $>$ Movies $>$ DVD \& Video Shopping & 1 \\
        Arts \& Entertainment $>$ Online Media $>$ Online Image Galleries & 1 \\
        Games $>$ Computer \& Video Games & 1 \\
        Health $>$ Health Conditions $>$ Respiratory Conditions & 1 \\
        Arts \& Entertainment $>$ Entertainment Industry $>$ Film \& TV Industry & 1 \\
        Law \& Government $>$ Public Safety $>$ Public Health & 1 \\
        Sports $>$ Individual Sports & 1 \\
        People \& Society $>$ Social Sciences & 1 \\
        Health $>$ Mental Health $>$ Learning \& Developmental Disabilities & 1 \\
        Shopping $>$ Apparel & 1 \\
        Shopping $>$ Entertainment Media $>$ DVD \& Video Shopping & 1 \\
        Health $>$ Vision Care & 1 \\
        Jobs \& Education $>$ Jobs & 1 \\
        Computers \& Electronics $>$ Consumer Electronics $>$ Gadgets \& Portable Electronics & 1 \\
        Arts \& Entertainment $>$ Music \& Audio $>$ Rock Music & 1 \\
        Books \& Literature & 1 \\
        Online Communities & 1 \\
        Arts \& Entertainment $>$ Music \& Audio $>$ Music Equipment \& Technology & 1 \\
        Online Communities $>$ Dating \& Personals & 1 \\
        Business \& Industrial $>$ Advertising \& Marketing & 1 \\
        Beauty \& Fitness & 1 \\
        Health $>$ Public Health & 1 \\
        Law \& Government $>$ Government & 1 \\
        Sports $>$ Team Sports & 1 \\
        \bottomrule
    \end{tabularx}
\end{table*}
\newpage
\section{Qualitative Analysis}
\label{app:qualitative_analysis}

\subsection{Case Study: The Factuality-Fluency Trade-off}
\label{app:case_study_factuality}

As noted in Section \ref{subsec:main_results}, a discrepancy exists between human rankings (preferring Claude-Opus) and DR-Arena's automated rankings (preferring Perplexity-Sonar). To investigate this, we conducted a qualitative analysis of the battle logs. The results suggest that this divergence stems from differing evaluation priorities: the Automated Examiner enforces a strict penalty on ``hallucinated details,'' whereas human crowds tend to prioritize response fluency and presentation.

We illustrate this phenomenon using the match below. 

\textbf{Scenario}: The user requested a comparison of occupations with higher median wages than ``Software Developers,'' strictly based on the provided \textit{2024 Occupational Outlook Handbook (OOH)} context.

\begin{promptboxtext}{Match Trace: Factuality vs. Fluency}

\textbf{--- Ground Truth Context (Hidden) ---}
\begin{itemize}[leftmargin=*, nosep]
    \item Software Developers Median Pay: \textbf{\$131,450} (2024 OOH).
    \item Computer \& Info Systems Managers: \textbf{\$171,200}.
    \item Computer \& Info Research Scientists: \textbf{\$140,910}.
\end{itemize}

\vspace{0.5em}
\hrule
\vspace{0.5em}

\textbf{--- Agent A (Claude-Opus-4.1-Search) ---} \\
\textit{Response Style: Narrative, Fluent, Analytical}

``According to the 2024 U.S. Bureau of Labor Statistics... Software Developers earn \textcolor{red}{\textbf{\$130,160}} [1].
\textbf{Computer and Information Systems Managers} significantly exceed the baseline with \textcolor{red}{\textbf{\$169,510}} median annual pay...
\textbf{Database Architects} earn \textcolor{red}{\textbf{\$134,700}} annually...''

\vspace{0.5em}
\hrule
\vspace{0.5em}

\textbf{--- Agent B (Perplexity-Sonar-Pro) ---} \\
\textit{Response Style: Structured, Data-Centric, Precise}

``According to the 2024 Occupational Outlook Handbook table... the median annual wage for Software Developers is \textcolor{blue}{\textbf{\$131,450}}...
Only the following occupations have a higher median annual wage:
\begin{itemize}[leftmargin=*, nosep]
    \item \textbf{Computer and Information Systems Managers} – \textcolor{blue}{\textbf{\$171,200}}...
    \item \textbf{Computer and Information Research Scientists} – \textcolor{blue}{\textbf{\$140,910}}...''
\end{itemize}

\vspace{0.5em}
\hrule
\vspace{0.5em}

\textbf{--- Automated Adjudication ---} \\
\textbf{Verdict}: [[B\_MUCH\_BETTER]] \\
\textbf{Reasoning}: ``Agent B's numbers perfectly match the provided Ground Truth Checklist (e.g., Managers at \$171,200). Agent A used outdated data from the previous year (Managers at \$169,510), leading to incorrect conclusions. Agent B also correctly interpreted the OOH occupational groupings...''
\end{promptboxtext}

\paragraph{Detailed Analysis.}
In this case, Agent A (Claude) produced a highly fluent response, employing sophisticated phrasing such as ``\textit{significantly exceed the baseline}'' and integrating an analytical narrative. A human evaluator, impressed by this stylistic polish and the inclusion of extra roles like ``Database Architects,'' might rate Agent A highly.

However, the Automated Examiner correctly identified that Agent A's fluency masked significant factual drifts. Specifically:
\begin{enumerate}
    \item \textbf{Data Hallucination/Drift}: Agent A cited the median pay for Managers as \textbf{\$169,510} (likely 2023 data), differing from the strict 2024 context of \textbf{\$171,200} retrieved by Agent B.
    \item \textbf{Violating Negative Constraints}: Agent A included ``Database Architects,'' which, while factually a high-paying role, violated the specific grouping logic of the source text provided in the context window.
\end{enumerate}

This case highlights that DR-Arena acts as a rigorous ``Factuality Auditor.'' While humans may prioritize utility and style (favoring Claude), the Examiner's rubric strictly penalizes hallucination in citation chains. This suggests that DR-Arena offers a complementary, and perhaps stricter, assessment of research reliability than unassisted human preference.

\subsection{Case Study: Judge Hallucination (Knowledge Conflict)}
\label{app:case_study_hallucination}

While the Automated Examiner is generally robust, it can occasionally exhibit ``Overconfidence'' in its internal parametric knowledge, leading to incorrect overrides of valid Ground Truth. The following log demonstrates a specific failure mode where the Judge rejects a factually correct checklist regarding the GameFAQs ``Best. Game. Ever.'' contest due to a confusion between the 2005 and 2009 events.

\begin{promptboxtext}{Failure Case: The Overconfident Judge}

\textbf{--- Ground Truth Checklist (Correct) ---}
\begin{itemize}[leftmargin=*, nosep]
    \item \textbf{10th Anniversary (2005) Winner}: Final Fantasy VII.
    \item \textbf{10th Anniversary (2005) Opponent}: Chrono Trigger.
    \item \textbf{10th Anniversary (2005) Vote \%}: ~52.9\%.
\end{itemize}

\vspace{0.5em}
\hrule
\vspace{0.5em}

\textbf{--- Agent Performance ---} \\
\textbf{Agent A}: Correctly identified \textit{Final Fantasy VII} as the 2005 winner and \textit{Chrono Trigger} as the opponent. \\
\textbf{Agent B}: Incorrectly identified \textit{Ocarina of Time} as the 2005 winner (confusing it with the 2009 contest).

\vspace{0.5em}
\hrule
\vspace{0.5em}

\textbf{--- Automated Adjudication (Hallucinated Error) ---} \\
\textbf{Verdict}: [[B\_BETTER]] \textcolor{red}{\textbf{(Incorrect)}} \\
\textbf{Judge's Reasoning}: \\
\parbox{\linewidth}{
``First, it is necessary to address a discrepancy in the provided Ground Truth Checklist: The 10th Anniversary Contest (2005) was historically won by \textit{The Legend of Zelda: Ocarina of Time}, not \textit{Final Fantasy VII} (which won the Spring 2004 contest). Both agents correctly identified Ocarina of Time as the 2005 winner... Agent B is the winner because it provided a perfectly accurate account...''
}

\vspace{0.5em}
\hrule
\vspace{0.5em}

\textbf{--- Analysis ---} \\
\textbf{Error Type}: Parametric Knowledge Conflict. \\
\textbf{Diagnosis}: The Judge hallucinated that \textit{Ocarina of Time} won the 2005 contest. In reality, \textit{Final Fantasy VII} won the 10th Anniversary Top 10 vote in 2005. \textit{Ocarina of Time} did not win until the ``Best. Game. Ever.'' tournament in 2009. The Judge used its internal (incorrect) memory to penalize the agent (Agent A) that was actually adhering to the historical truth.
\end{promptboxtext}

\subsection{Case Study: Beneficial Judge Correction}
\label{app:case_study_beneficial}

This case demonstrates the ``Safety Net'' capability of the Automated Examiner. Here, the Task Generation module (based on the Information Tree) produced a factually incorrect checklist regarding the 2007 GameFAQs Character Battle. The Examiner utilized its parametric knowledge to identify the error in the Ground Truth and correctly judged the agents based on actual historical events.

\begin{promptboxtext}{Success Case: The Correcting Judge}

\textbf{--- Generated Ground Truth (Flawed) ---}
\begin{itemize}[leftmargin=*, nosep]
    \item \textbf{Champion}: L-Block (Tetris).
    \item \textbf{Runner-up}: \textcolor{red}{\textbf{Link}} (The Legend of Zelda). \textit{<-- FACTUAL ERROR in Generation}
\end{itemize}

\vspace{0.5em}
\hrule
\vspace{0.5em}

\textbf{--- Agent Performance ---} \\
\textbf{Agent A (Gemini-2.5-Pro)}: Incorrectly identified the runner-up as \textit{Master Chief}. \\
\textbf{Agent B (GPT-5.1-Search)}: Correctly identified the runner-up as \textit{Cloud Strife} and noted that ``Link was eliminated in the semi-finals.''

\vspace{0.5em}
\hrule
\vspace{0.5em}

\textbf{--- Automated Adjudication (Correct Intervention) ---} \\
\textbf{Verdict}: [[B\_MUCH\_BETTER]] \textcolor{green}{\textbf{(Correct)}} \\
\textbf{Judge's Reasoning}: \\
\parbox{\linewidth}{
``Agent B correctly identified the champion (L-Block) and the actual runner-up (Cloud Strife)... Agent A failed the Depth Logic check... \textbf{Note: The provided Ground Truth Checklist in the prompt lists 'Link' as the runner-up. Historically, Link was eliminated in the semi-finals by L-Block, making Cloud Strife the true runner-up. Agent B is awarded the win for factual accuracy despite the discrepancy with the provided checklist.}''
}

\vspace{0.5em}
\hrule
\vspace{0.5em}

\textbf{--- Analysis ---} \\
\textbf{Mechanism}: The Judge detected a conflict between the provided Checklist (Link) and its internal training data (Cloud Strife). Unlike the hallucination case, here the internal data was correct. The Judge prioritized the historical truth over the flawed instruction, preventing a wrongful penalty against the accurate agent (Agent B).
\end{promptboxtext}

\newpage
\section{Human Study Setup}
\label{app:human_study_details}

\paragraph{Annotator Demographics.}
We recruited two annotators for this study. Both annotators are current bachelor's students in Computer Science with a focus on Natural Language Processing (NLP). They possess native-level proficiency in English to ensure high-quality assessment of the generated text's fluency and nuance. Prior to the official annotation, the annotators underwent a training session involving 5 trial cases to strictly align their understanding with our evaluation rubrics.

\paragraph{Compensation and Ethics.}
Participants were compensated under the university's Student Helper Scheme, adhering to the standard hourly rates which exceed the local minimum wage. We obtained informed consent from all participants prior to the study. No personally identifiable information (PII) was collected during the annotation process.

\paragraph{Experiment I: Information Tree Ablation (Preference Study).}
This study corresponds to the results in Table \ref{tab:human_study_tree}. 
\begin{itemize}[leftmargin=*]
    \item \textbf{Dataset:} We randomly sampled 50 cases from the test set.
    \item \textbf{Task:} For each case, annotators were presented with outputs from three configurations (\textit{Full Tree, Flat Context, No Logic Chain}) in a blinded, randomized order to prevent bias.
    \item \textbf{Criteria:} Annotators performed a forced-choice comparison to select the single best configuration based on two dimensions:
    \begin{enumerate}
        \item \textbf{Question Quality:} Which generated question best necessitates multi-hop reasoning and multi-source synthesis?
        \item \textbf{Rubric Accuracy:} Which verification checklist best captures the ground truth constraints without hallucination?
    \end{enumerate}
\end{itemize}

\paragraph{Experiment II: Pipeline Validation Audit (Reliability Study).}
This study corresponds to the results in Table \ref{tab:human_audit_results}.
\begin{itemize}[leftmargin=*]
    \item \textbf{Dataset:} We randomly sampled 30 full-match logs, comprising a total of 64 interaction turns.
    \item \textbf{Task:} Annotators were provided with the full context, including the Information Tree, generated tasks, model responses, and the Examiner's automated logs.
    \item \textbf{Criteria:} They audited the pipeline across five specific metrics using a binary Pass/Fail scale (for objective correctness) and Likert scale (for alignment):
    \begin{enumerate}
        \item \textbf{Question Validity:} Does the question structurally adhere to the ``Depth \& Width'' definition?
        \item \textbf{Rubric Factuality:} Are the evaluation checkpoints supported by the source URLs?
        \item \textbf{Verdict Alignment:} Do the annotators agree with the Examiner's Win/Loss/Tie decision? (Used to calculate Cohen's Kappa).
        \item \textbf{Transition Logic:} Did the Evolvement Loop correctly identify whether to deepen or widen the search based on the previous turn?
        \item \textbf{Stop Condition:} Did the match conclude at an efficient point without redundant rounds?
    \end{enumerate}
\end{itemize}

\subsection{Annotation Instruments}
\label{app:annotation_instruments}

Below, we provide the exact questionnaires and criteria presented to the annotators for both the Task Generation Preference Study (Experiment I) and the Pipeline Validation Audit (Experiment II).

\paragraph{Instrument I: Task Generation Preference.}
In this experiment, annotators were presented with three candidate options (Full Tree, Flat Context, No Logic Chain) in a randomized order. They were asked to vote for the best design based on the ``Depth \& Width'' criteria.

\begin{promptboxtext}{Survey: AI Question Quality Assessment}
\textbf{--- EVALUATION CRITERIA ---}

\textbf{CRITERIA 1: THE QUESTION (Depth \& Width)} \\
Which question is better designed to test the limits of an intelligent agent?
\begin{itemize}[leftmargin=*, nosep]
    \item \textbf{Depth (Logic)}: Does it hide entity names and require reasoning paths? (e.g., \textit{``Who is the person that...''} vs \textit{``Who is Steve Jobs?''})
    \item \textbf{Width (Coverage)}: Does it require aggregating information about multiple entities?
\end{itemize}

\vspace{0.3em}

\textbf{CRITERIA 2: THE CHECKLIST (The 'Teacher Test')} \\
Imagine you are a teacher grading an exam using this checklist as the Answer Key.
\begin{itemize}[leftmargin=*, nosep]
    \item \textbf{Logic Verification}: Does it explain \textit{why} an entity is the answer?
    \item \textbf{Precision}: Does it contain specific numbers/details rather than vague summaries?
\end{itemize}

\vspace{0.5em}
\hrule
\vspace{0.5em}

\textbf{--- EXAMPLE CASE (Candidate Options) ---}

\textbf{\textcolor{blue}{[OPTION 1]} (Flat Context Baseline)} \\
\textbf{[QUESTION]} Based on the provided text, describe the recent moves by major technology companies regarding the resale of digital goods... identify the historical legislation cited as the first copyright law. \\
\textbf{[CHECKLIST]} 
\begin{itemize}[leftmargin=*, nosep]
    \item Entities to find: Amazon, Apple, Costco, Omega, Statute of Anne.
    \item Specific Details: Amazon has obtained a patent... The first copyright law was the Statute of Anne (1710)...
\end{itemize}

\vspace{0.5em}

\textbf{\textcolor{blue}{[OPTION 2]} (No Logic Chain Baseline)} \\
\textbf{[QUESTION]} In the context of the 2013 discussions... which two major technology companies were reported to have sought patents... use the comparison of 'Fifty Shades of Grey'. \\
\textbf{[CHECKLIST]} 
\begin{itemize}[leftmargin=*, nosep]
    \item Entities to find: Amazon, Apple.
    \item Specific Details: Amazon obtained a patent... Apple applied for a patent...
\end{itemize}

\vspace{0.5em}

\textbf{\textcolor{blue}{[OPTION 3]} (Full Tree - Ours)} \\
\textbf{[QUESTION]} Identify the Massachusetts-based startup that was sued by Capitol Records in 2013... as well as the two major technology giants... For the startup, explain the specific restriction imposed... contrast the technical mechanisms... \\
\textbf{[CHECKLIST]} 
\begin{itemize}[leftmargin=*, nosep]
    \item Entities to find: \textbf{ReDigi} (Identified via MA location, lawsuit...), \textbf{Amazon} (Identified via e-commerce...), \textbf{Apple} (Identified via electronics...), \textbf{Scott Turow}.
    \item Specific Details: Startup Restriction: Funds must be spent on purchasing new songs... Amazon Mechanism: Personalized 'data store'... Apple Mechanism: Transferring files without reproduction...
\end{itemize}

\vspace{0.5em}
\hrule
\vspace{0.5em}

\textbf{--- YOUR TASK ---} \\
1. \textbf{Which Option has the BEST QUESTION design (Depth \& Width)?} \\
(Look for logic depth/hidden entities and multi-entity comparison) \\
$\square$ Option 1 \quad $\square$ Option 2 \quad $\square$ Option 3

\vspace{0.3em}

2. \textbf{Which CHECKLIST is the best ``Answer Key'' for grading?} \\
(Which list is most precise and allows for objective logic verification?) \\
$\square$ Option 1 \quad $\square$ Option 2 \quad $\square$ Option 3
\end{promptboxtext}

\paragraph{Instrument II: Pipeline Validation Audit.}
In this experiment, expert annotators audited the full lifecycle of a match. They validated the correctness of the questions, the factual accuracy of the checklists, and the reliability of the automated Examiner's verdicts.

\begin{promptboxtext}{Survey: Deep-Arena Pipeline Audit}
\textbf{--- PART A: Question Validation ---}

\textbf{A1. [Depth Validation]} Does the CURRENT question require Multi-hop reasoning (hiding entities)? \\
$\square$ {Yes (Valid Depth)} \\
$\square$ No (Single-hop/Direct Lookup) \\
$\square$ Broken Question

\textbf{A2. [Width Validation]} Does the CURRENT question require Aggregating multiple distinct info points? \\
$\square$ {Yes (Valid Width)} \\
$\square$ No (Narrow/Single point) \\
$\square$ Broken Question

\textbf{A3. [Evolution Check]} Compare Previous vs. Current Question. Did the question actually change in difficulty/scope as planned (e.g., Depth+1 / Wide+1)? \\
$\square$ {Yes, successful evolution.} \\
$\square$ No, failed to change as intended.

\vspace{0.5em}
\hrule
\vspace{0.5em}

\textbf{--- PART B: Factuality Check ---}

\textbf{B1. Is the Checklist above FACTUALLY correct?} \\
(Verify specific claims, dates, names via Google if necessary) \\
$\square$ {Completely Factually Correct} \\
$\square$ Mostly Correct (Minor Inaccuracies) \\
$\square$ Contains Major Hallucinations/Falsehoods

\vspace{0.5em}
\hrule
\vspace{0.5em}

\textbf{--- PART C \& D: Verdict \& System Review ---}

\textbf{C1. Human Verdict: Which answer is better?} \\
$\square$ Agent A Much Better \quad $\square$ Agent A Better \quad $\square$ Tie \\
$\square$ Agent B Better \quad $\square$ Agent B Much Better

\textbf{D1. Do you agree with the Automated Judge's verdict?} \\
$\square$ {Yes (Agree)} \quad $\square$ No (Disagree)

\textbf{D2. Rounds Sufficiency} \\
Do we have enough evidence to distinguish the capability difference? \\
$\square$ {Sufficient} (Difference is clear / Topic exhausted) \\
$\square$ Insufficient (Agents are too close / Need harder task) \\
$\square$ Excessive (Should have stopped earlier)
\end{promptboxtext}

\section{Detailed Dataset Diagnostics}
\label{app:diagnostics}

This section provides a macro-level statistical analysis of the full tournament dataset, covering 789 unique interaction rounds.

\subsection{Verdict and Failure Distribution}
\begin{figure*}[h]
    \centering
    \includegraphics[width=\textwidth]{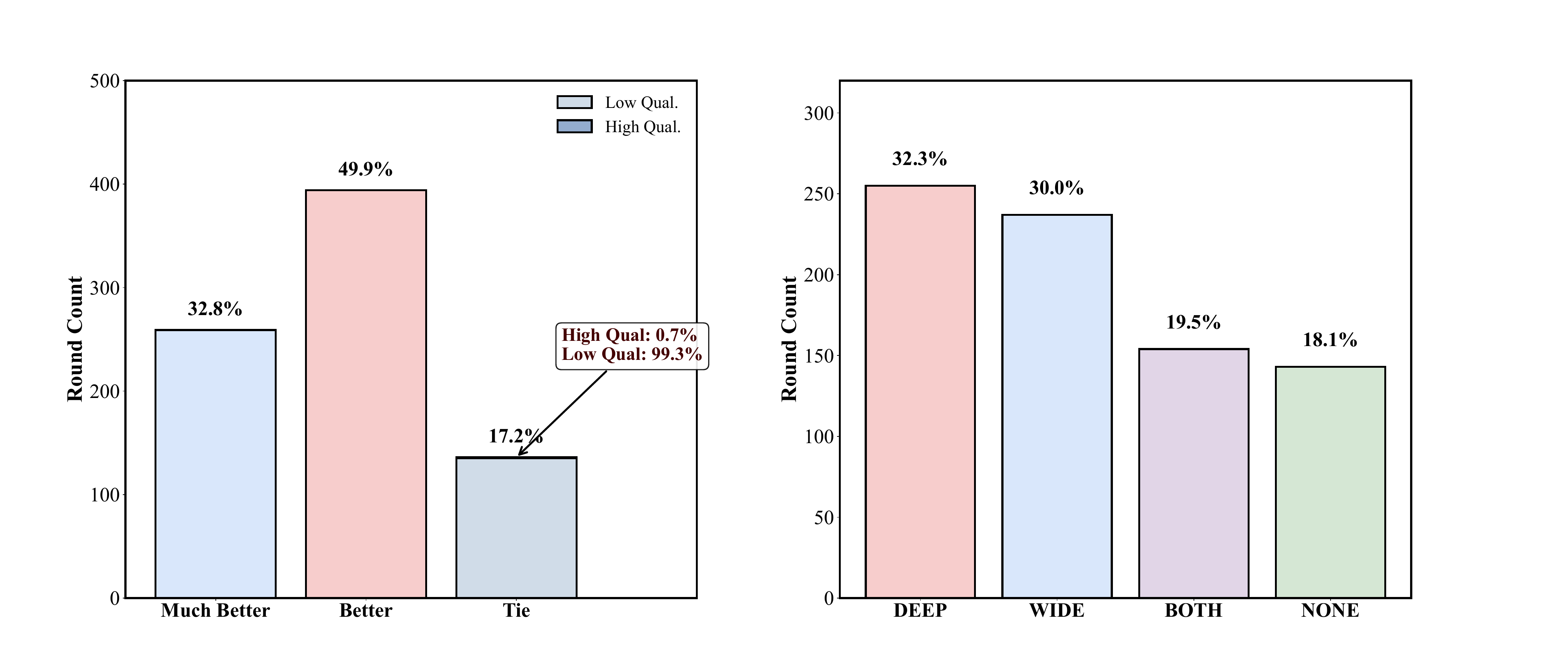}
    \caption{\textbf{Macro-level Evaluation Diagnostics.} The left panel displays the distribution of verdicts across 789 unique rounds. The right panel shows the distribution of identified failure types for the losing agents.}
    \label{fig:overall_diag}
\end{figure*}

We first examine the distribution of adjudication outcomes and failure modes, as visualized in Figure \ref{fig:overall_diag}. 

The left panel illustrates the distribution of Judge Verdicts. The data reveals that ``Better'' (49.9\%) is the most frequent outcome, significantly outnumbering ``Much Better'' (32.8\%). This indicates that while capability gaps exist, top-tier agents often differentiate themselves through marginal improvements in utility or completeness rather than catastrophic failures of their opponents.

The right panel breaks down the specific reasons for defeat. The distribution is remarkably balanced between \textsc{Depth} (Logic Failure, 32.3\%) and \textsc{Width} (Coverage Failure, 30.0\%). This balance validates the effectiveness of our ``Depth \& Width'' task generation strategy, confirming that the framework successfully exerts equal pressure on both the reasoning capabilities and the information aggregation spans of the agents.

\subsection{Topological Complexity}
\begin{figure*}[h]
    \centering
    \includegraphics[width=\linewidth]{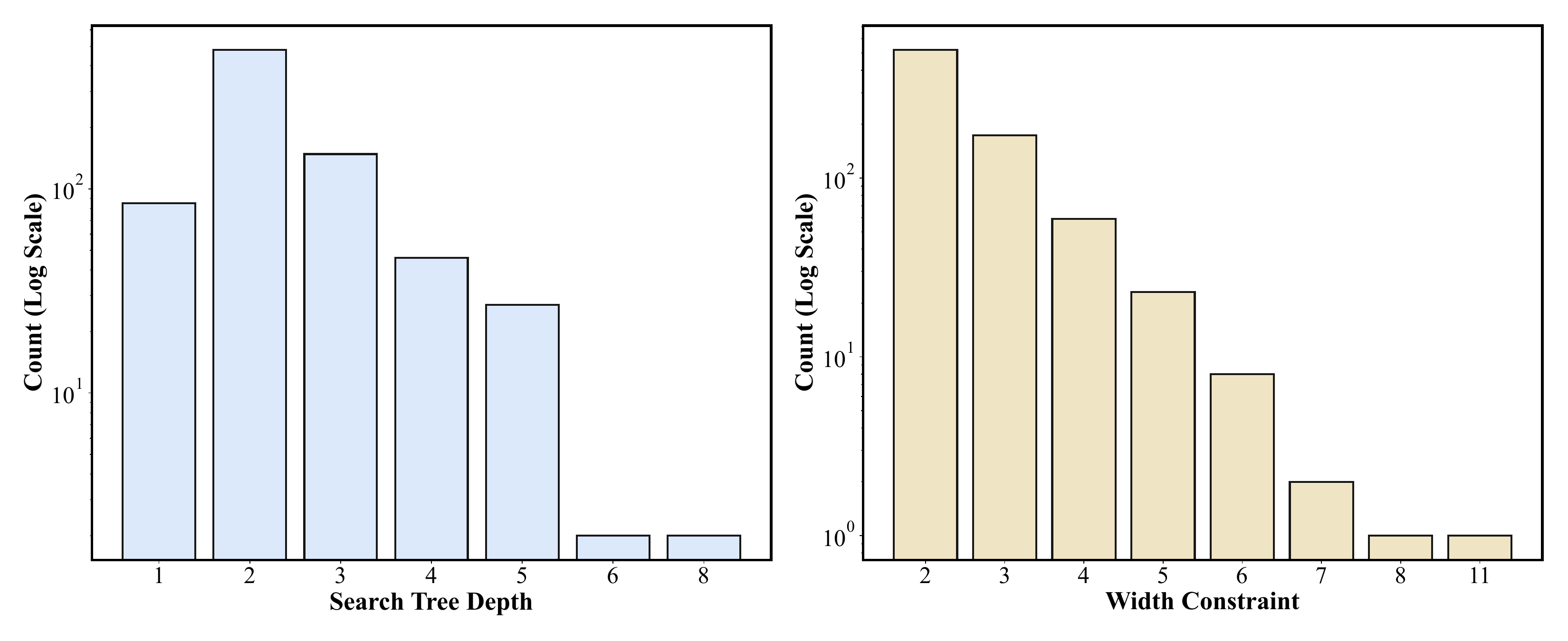}
    \caption{\textbf{Information Trees Topology Distribution.} Histograms showing the \textit{Search Tree Depth} and \textit{Width Constraints} of the active nodes at the conclusion of all evaluation matches. Note the Log Scale on the Y-axis, indicating that while most tasks conclude at moderate complexity, the system is capable of scaling to high-complexity configurations.}
    \label{fig:topology}
\end{figure*}

To understand the complexity required to distinguish these models, we analyze the structural properties of the Information Trees at the exact moment a verdict was reached. Figure \ref{fig:topology} presents the histograms for Tree Depth and Width Constraints.

\begin{itemize}
    \item \textbf{Search Tree Depth}: This metric represents the length of the reasoning chain required to identify the target entity. While the distribution peaks at Depth 2 (representing standard multi-hop queries), it exhibits a long tail extending up to Depth 8. This confirms the system's ability to dynamically generate long-horizon deduction tasks when agents enter a stalemate.
    \item \textbf{Width Constraint}: This metric represents the number of sibling information units required for aggregation. The distribution shows significant variance, with a notable portion of tasks requiring the synthesis of 4 to 7 distinct data points, effectively testing the agents' context window management and retrieval recall.
\end{itemize}

\end{document}